\documentclass{article}

\usepackage{wrapfig}
\usepackage{lipsum}
\usepackage{microtype}
\usepackage{graphicx}
\usepackage{subcaption}
\usepackage{booktabs} %
\usepackage{enumitem}

\usepackage{hyperref}

\usepackage[accepted]{icml2023}

\usepackage{amsmath}
\usepackage{amssymb}
\usepackage{mathtools}
\usepackage{amsthm}

\usepackage[capitalize,noabbrev]{cleveref}

\usepackage{minitoc}

\usepackage{rotating}

\usepackage{tikz}
\usetikzlibrary{positioning}

\theoremstyle{plain}
\newtheorem{theorem}{Theorem}[section]

\theoremstyle{definition}
\newtheorem{definition}[theorem]{Definition}

\theoremstyle{remark}

\newcommand{\cG}{\mathcal{G}}
\newcommand{\cS}{\mathcal{S}}
\newcommand{\cC}{\mathcal{C}}
\newcommand{\cB}{\mathcal{B}}
\newcommand{\cP}{\mathcal{P}}

\newcommand{\cV}{\mathcal{V}}
\newcommand{\cX}{\mathcal{X}}

\newcommand{\HighConf}{\textsc{Sure}}
\newcommand{\LowConf}{\textsc{Unsure}}

\newcommand{\RUSURE}{\textsc{R-U-SURE}}

\newcommand{\RUSUREPREFIX}{\textsc{R-U-SURE} (Prefix)}
\newcommand{\RUSUREPREFIXPLUSUNSURE}{\textsc{R-U-SURE} (Prefix+Region)}
\newcommand{\RUSUREREGIONS}{\textsc{R-U-SURE} (Region)}
\newcommand{\RUSUREAPI}{\textsc{R-U-SURE} (API)}

\newcommand\MAXNUMBEROFMODELSAMPLES{31}

\newcommand{\smminus}{\scalebox{0.6}[1.0]{$-$}}
\definecolor{insertcolor}{HTML}{268bd2}
\definecolor{deletecolor}{HTML}{dc322f}
\definecolor{matchcolor}{HTML}{859900}
\definecolor{partitioncolor}{HTML}{6c71c4}

\usepackage{amsmath,amsfonts,bm}

\def\eqref#1{equation~\ref{#1}}

\def\1{\bm{1}}

\def\vb{{\bm{b}}}

\def\vlambda{{\bm{\lambda}}}

\DeclareMathAlphabet{\mathsfit}{\encodingdefault}{\sfdefault}{m}{sl}
\SetMathAlphabet{\mathsfit}{bold}{\encodingdefault}{\sfdefault}{bx}{n}

\newcommand{\E}{\mathbb{E}}

\newcommand{\R}{\mathbb{R}}

\DeclareMathOperator*{\argmax}{arg\,max}
\DeclareMathOperator*{\argmin}{arg\,min}

\usepackage{makecell}

\definecolor{commentcolor}{gray}{0.6}

\usepackage[textsize=tiny]{todonotes}

\icmltitlerunning{R-U-SURE\hfill\thepage}

\begin{document}

\twocolumn[
\icmltitle{R-U-SURE? Uncertainty-Aware Code Suggestions \\ By Maximizing Utility Across Random User Intents}

\begin{icmlauthorlist}
\icmlauthor{Daniel D. Johnson}{google,toronto}
\icmlauthor{Daniel Tarlow}{google}
\icmlauthor{Christian Walder}{google}
\end{icmlauthorlist}

\icmlaffiliation{google}{Google Research, Brain Team}
\icmlaffiliation{toronto}{University of Toronto, Department of Computer Science}

\icmlcorrespondingauthor{Daniel D. Johnson}{ddjohnson@google.com}

\icmlkeywords{Machine Learning, ICML}

\vskip 0.3in
]

\printAffiliationsAndNotice{}  %

\begin{abstract}
Large language models show impressive results at predicting structured text such as code, but also commonly introduce errors and hallucinations in their output. When used to assist software developers, these models may make mistakes that users must go back and fix, or worse, introduce subtle bugs that users may miss entirely. We propose \emph{Randomized Utility-driven Synthesis of Uncertain REgions (\mbox{R-U-SURE})}, an approach for building uncertainty-aware suggestions based on a decision-theoretic model of goal-conditioned utility, using random samples from a generative model as a proxy for the unobserved possible intents of the end user. Our technique combines minimum-Bayes-risk decoding, dual decomposition, and decision diagrams in order to efficiently produce structured uncertainty summaries, given only sample access to an arbitrary generative model of code and an optional AST parser. We demonstrate R-U-SURE on three developer-assistance tasks, and show that it can be applied different user interaction patterns without retraining the model and leads to more accurate uncertainty estimates than token-probability baselines.
We also release our implementation as an open-source library at \url{https://github.com/google-research/r_u_sure}.

\end{abstract}

\section{Introduction}\label{sec:intro}

\begin{figure}[t!]
    \centering

\fbox{
\begin{minipage}{0.95\linewidth} %
\includegraphics[width=\linewidth,page=2,trim=2.05in 3.23in 2.05in 1.22in, clip]{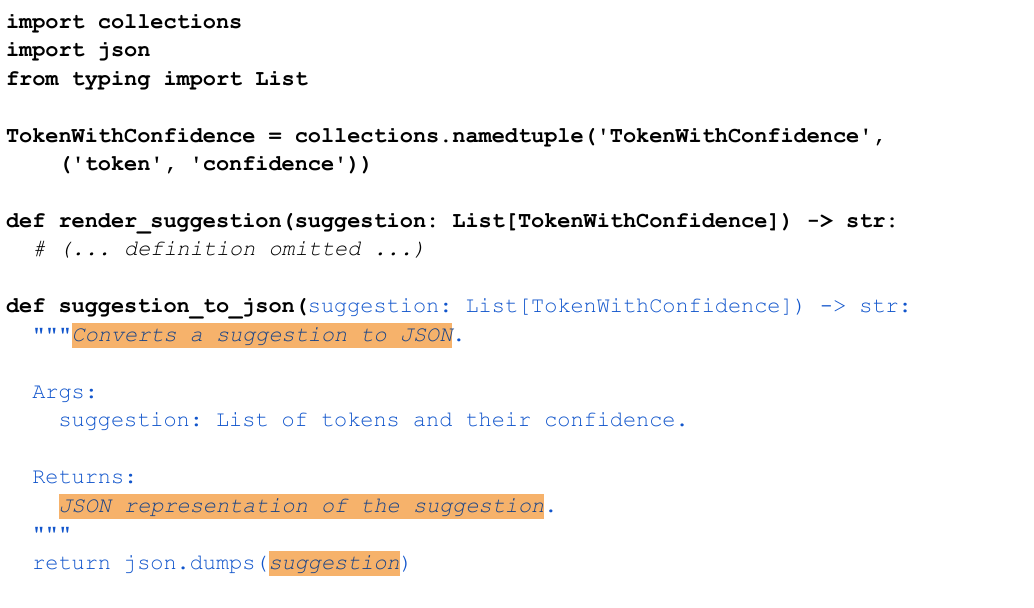}
\end{minipage}
}
\\[0.01\linewidth]
\fbox{
\includegraphics[width=0.44\linewidth,page=3,trim=0.86in 2.43in 5.56in 1.44in, clip]{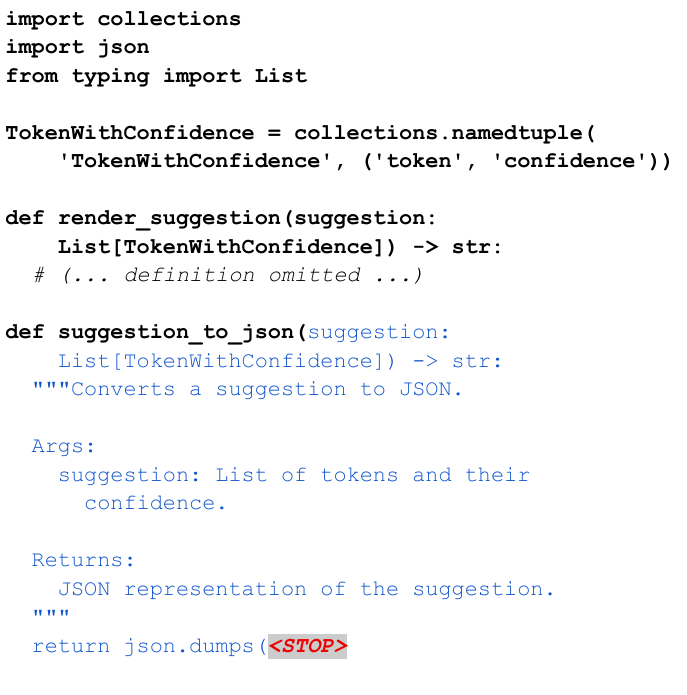}
}
\null\hfill
\fbox{
\includegraphics[width=0.44\linewidth,page=3,trim=5.54in 2.43in 0.88in 1.44in, clip]{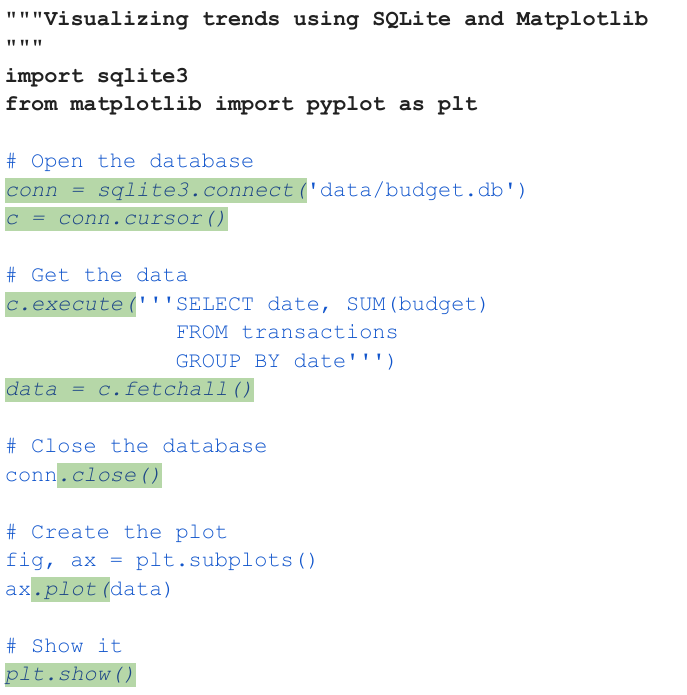}
}
    
    \caption{
    Given a partial file as context (bolded black code) and outputs from a fixed language model (blue code), our approach can be used to predict parts of generated programs that may need editing (top, orange background), adjust completion length to avoid uncertain parts (left, red text), or identify the most relevant statements from a larger prediction (right, green background), by searching over a space of uncertainty-augmented suggestions $\cS$. (Examples lightly reformatted to fit this figure; see \cref{app:example-outputs}.)
    }
    \label{fig:frontpage}
\end{figure}
\begin{figure*}
    \centering
    \includegraphics[width=1\linewidth,trim=0 16.6in 2.75in -0.2in,clip]{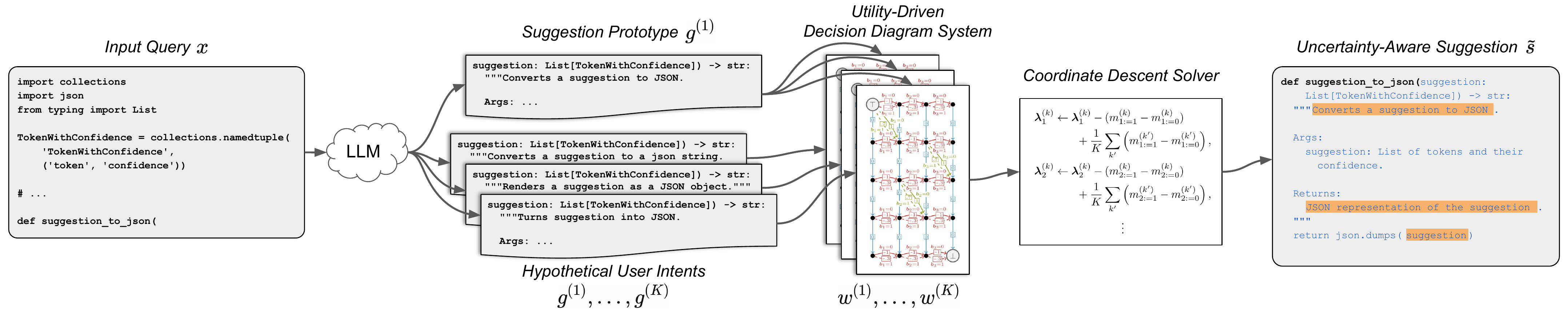}
    
    \caption{
    Overview of the R-U-SURE system. 
    Given an input context $x$, we generate a set of samples $g^{(1)},\dots,g^{(K)}$ from a pretrained generative model. We select one sample $g^{(1)}$ as a \emph{suggestion prototype} and interpret the full set of samples as hypothetical user intents (e.g. examples of code the user may want to write), as described in \cref{sec:approx-true-intents-model-samples}. For each hypothetical intent $g^{(k)}$, we can estimate the utility $u(g^{(k)}, s)$ of showing some concrete suggestion $s$ to a user who intended to write $g^{(k)}$ (discussed in \cref{sec:problem-statement}). We represent this efficiently using a system of decision diagrams $w^{(1)},\dots,w^{(k)}$ (\cref{sec:representing-as-decision-diagrams}), and use coordinate descent (\cref{subsec:dual-decomposition-decision-diagrams}) to identify a set of annotations to insert into the suggestion prototype $g^{(1)}$ to maximize expected utility over these samples, obtaining an uncertainty-aware suggestion $\tilde{s}$ that is likely to be more useful for the user's unobserved true intent.
    }
    \label{fig:overview}
\end{figure*}

Large language models have demonstrated remarkable abilities for generating both natural language \citep{Brown2020LanguageMAGPT,Chowdhery2022PaLMSL} and source code
\citep{svyatkovskiy2020intellicode,Feng2020CodeBERTAP,Chen2021EvaluatingLLCodex,li2022alphacode,Nijkamp2022CodeGenAO}. These abilities have led them to be incorporated into a number of developer assistance tools and services, such as \href{https://github.com/features/copilot}{GitHub Copilot} and \href{https://www.tabnine.com/}{Tabnine}.

Unfortunately, when faced with novel or unpredictable situations, large language models have a tendency to guess or ``hallucinate'' unwanted outputs
\citep{Maynez2020OnFA,Liu2021ATR}. In the context of software development, these guesses can slow development by requiring developers to spend time verifying the suggestion and deleting any incorrect parts  \citep{Mozannar2022ReadingBT,Barke2022GroundedCH,Upadhyaya2022ExpectationVE}, or worse, lead to undetected problems and less secure code \citep{Pearce2021AnEC}.
Compounding this issue is the presence of \emph{automation bias}, an effect where users fail to notice issues in outputs of automated systems  \citep{Madi2022HowRI,Cummings2004AutomationBI,Lyell2017AutomationBA}.

An interesting property of generated code suggestions is that some parts of the user's intent (e.g. control flow and API boilerplate) can be predicted more easily than others (e.g. edge-case behavior or the signatures of novel functions). User interaction studies of ML coding assistants have revealed that software engineers would benefit if suggestions included indicators of model uncertainty \citep{Mozannar2022ReadingBT,Weisz2021PerfectionNR} or user-fillable ``holes'' \citep{Barke2022GroundedCH}. However, \citet{Vasconcelos2022GenerationPA} have found that per-token conditional probability estimates are insufficient to provide good predictions of necessary edits. \citet{Guo2021Grammformer} propose a top-down generative model of code that uses a programming language's grammar to generate completions containing holes, but these holes must align with grammar nonterminals and cannot identify uncertain subregions within lists of statements or expressions.

In this work, we show that there is a way to harness the remarkable capabilities of pretrained language models to both generate high-quality code suggestions and also produce concise and semantically-meaningful representations of their own uncertainty, without requiring any fine-tuning. Our key insight is that, since language models of code are trained to predict file contents from context, we can reinterpret the samples from a well-trained language model as \emph{plausible goal states} for the user. We can then use these samples to estimate how useful our suggestions would be for a user whose intent we do not know, and to \emph{modify} those suggestions to make them useful across a diverse set of possible user intents.

As a concrete motivating example, consider the task of code completion under uncertainty, and suppose we wish to highlight specific regions of a completion suggestion to help end-users identify the parts of the suggestion they need to change, as shown at the top of \cref{fig:frontpage}. To do this, we can define a space $\cS$ of annotated suggestions, where some tokens are highlighted as \LowConf{}. We can then approximate how helpful a suggestion $s \in \cS$ would be to a user who actually wants to write code $g$ by computing a confidence-adjusted edit distance between $s$ and $g$, assuming that \LowConf{} tokens will be double-checked by the user and thus be easier to edit if wrong but also save less time (and thus be less useful) than non-highlighted \HighConf{} tokens if they turn out to be correct. If we can find a set of annotations that has high utility across many hypothetical goals $g^{(k)}$ drawn from a language model, then as long as the language model is well calibrated, the \LowConf{} annotations should provide a summary of the model's uncertainty that is similarly useful for accomplishing the user's unknown goal $g$.

Our contributions are as follows:
\begin{itemize}%
    \item We describe a utility-driven framework (summarized in \cref{fig:overview}) for producing uncertainty-aware suggestions given only sample-access to an arbitrary language model, by interpreting its samples as plausible user intents and using combinatorial optimization to identify the highest-utility suggestion, extending sample-based minimum Bayes risk decoding \citep{eikema2020bayes-risk-decoding}.
    \item We show how to apply dual decomposition
    to a novel decision diagram representation of edit-distance-based utility functions, yielding an efficient coordinate-descent optimizer and building a bridge between recent advances in language model decoding and combinatorial optimization.
    \item We construct a number of variants for our utility functions, enabling them to incorporate tree structure from an error-tolerant syntax-tree parser, account for both deletions and insertions, and respond to uncertainty by either annotating or removing the uncertain parts.
    \item We demonstrate our approach across three developer-assistance-inspired tasks (visualized in \cref{fig:frontpage}), and show that our approach yields a better tradeoff between correct and incorrect predictions than heuristics based on cumulative or per-token probability.
\end{itemize}

\section{Problem Statement}\label{sec:problem-statement}

We tackle the problem of providing \emph{contextual, uncertainty-aware suggestions} to assist users of ML-integrated tools with unobserved goals, with a particular focus on assisting software development. As we discuss in Section \ref{sec:intro}, there may not be enough information to fully determine the user's intent given the context.
Our strategy is thus to augment the space of possible suggestions to account for the uncertainty in the user's intent in an explicit way.
For instance, we can insert visual markers into a code-completion suggestion to draw attention to the parts of the suggestion that the user may wish to change. By doing so, we can avoid silently introducing incorrect behavior, and produce a suggestion that is useful regardless of what intent the user actually has.

We formalize this intuition using a decision theoretic framework. We let $\cX$ be a set of contexts (e.g. the current partially-written code file and any other relevant IDE state) and $\cG$ be a set of goals (e.g. the desired final state of the code file), with the specific context $x \in \cX$ and goal $g \in \cG$ of each user being distributed according to some unknown distribution $p(X=x, G=g)$.
We further specify a set $\cS$ of possible system suggestions along with a \emph{utility metric} $u : \cG \times \cS \to \R$, where $u(g,s)$ measures (or approximates) \emph{how useful} suggestion $s$ is toward accomplishing some specific goal $g$ that the user might have. 

Consider again the motivating example introduced in \cref{sec:intro}. Letting $\Sigma$ be a set of tokens, we can define $\cG = \Sigma^*$ as the set of possible token sequences $g = [g_1, g_2, \dots, g_M] \in \cG$ the user may wish to write, with each $g_i \in \Sigma$. We can then define $\cS = (\Sigma \times \cC)^*$ as a set containing uncertainty-annotated suggestions $s = [(s_1, c_1), (s_2, c_2), \dots, (s_N, c_N)] \in \cS$, where each suggestion is a sequence of pairs of tokens $s_i \in \Sigma$ and confidence indicators $c_i \in \cC = \{\HighConf{}, \LowConf{}\}$. Finally, we can define $u(g, s)$ based on the edit distance from $s$ to $g$, with a smaller penalty for deleting incorrect \LowConf{} tokens but a smaller reward for keeping correct ones. 
An example of how such a $u$ might be implemented using dynamic programming is shown in \cref{alg:simple-edit-distance-dp} of \cref{fig:decision-diagrams}, where $\alpha$  is the reward for keeping correct $\LowConf{}$ tokens and  $\beta$ is the penalty for deleting incorrect $\LowConf{}$ tokens. In \cref{sec:variants} we discuss how we extend this idea to account for program syntax trees and inserted code.

More generally, we can think about each $s \in S$ as a possible suggestion our system could show, and use $u$ to estimate the usefulness of that suggestion for a particular goal.
For a given context $x \sim p(X)$, we wish to find a concrete suggestion $s^*$ which is as useful as possible, e.g. that maximizes $u(g, s^*)$, in the presence of uncertainty about $g$. If we knew the true distribution $p(G|X)$, we might seek the suggestion that is most useful on average over
the user's likely intents:
\begin{equation}
s^* = \argmax_{s \in \cS} \E_{g \sim p(G|X = x)}[u(g, s)]
\label{eqn:true-distn-objective}
\end{equation}
This choice is also known as the \emph{minimum Bayes risk} suggestion, as it minimizes the expected risk (negative utility) of the action under the conditional distribution $p(G|X)$.

\section{Approach}
Unfortunately, we do not have access to the distribution in \cref{eqn:true-distn-objective}.
We now present
Randomized Utility-driven Synthesis of Uncertain REgions (\mbox{R-U-SURE}), a tractable procedure for approximating $s^*$ by combining samples from a model using combinatorial optimization.

\begin{figure*}[ht]
    \centering

        \scalebox{0.77}{
        \begin{minipage}{0.48\linewidth}
        \begin{algorithm}[H]\footnotesize
        \caption{Sequence edit-distance utility $u(g,s)$ \vphantom{$w^{(k)}$}}
        \label{alg:simple-edit-distance-dp}
        \begin{minipage}[c][20.25em][c]{\textwidth}
        \begin{algorithmic}
        \STATE {\bfseries Input:} $g = [g_1, \dots, g_M], s = [(s_1, c_1), \dots, (s_N, c_N)], \alpha, \beta$
        \STATE Initialize $T^i_j$ to $-\infty$ for $0 \le i \le M, 0 \le j \le N$
        \STATE $T^0_0 \gets 0.0$ \COMMENT{~~\# Base case for dynamic programming}
        \FOR{$i=0$ {\bfseries to} $M$, $j=0$ {\bfseries to} $N$}
            \IF[\textcolor{matchcolor}{~~\# Match $s_j$ and $g_i$}]{$i> 0$ \AND $j > 0$ \AND $g_i = s_j$}
                \STATE
                \textbf{if} $c_j = \HighConf{}$ \textbf{then} $T^{i}_{j} \gets \max(T^{i}_{j}, T^{i-1}_{j-1} + 1)$
                \STATE
                \textbf{if} $c_j = \LowConf{}$ \textbf{then} $T^{i}_{j} \gets \max(T^{i}_{j}, T^{i-1}_{j-1} + \alpha)$
            \ENDIF
            \IF[\textcolor{insertcolor}{~~\# Insert $g_i$}]{$i > 0$}
                \STATE $T^{i}_{j} \gets \max(T^{i}_{j}, T^{i-1}_{j} + 0)$
            \ENDIF
            \IF[\textcolor{deletecolor}{~~\# Delete $s_j$}]{$j > 0$}
                \STATE
                \textbf{if} $c_j = \HighConf{}$ \textbf{then} $T^{i}_{j} \gets \max(T^{i}_{j}, T^{i}_{j-1} - 1)$
                \STATE
                \textbf{if} $c_j = \LowConf{}$ \textbf{then} $T^{i}_{j} \gets \max(T^{i}_{j}, T^{i}_{j-1} - \beta)$
            \ENDIF
        \ENDFOR
        \RETURN $T^{M}_{N}$
        \end{algorithmic}
        \end{minipage}
        \end{algorithm}
        \end{minipage}
        }
\null\hfill
        \scalebox{0.77}{
        \begin{minipage}{0.49\linewidth}
        \begin{algorithm}[H]\footnotesize
        \caption{Decision diagram for $w^{(k)}(\vb)$}
        \label{alg:simple-edit-distance-decision-diagram}
        \begin{minipage}[c][20.25em][c]{\textwidth}
        \begin{algorithmic}
        \STATE {\bfseries Input:} $g^{(k)} = [g_1, g_2, \dots, g_M], g^{(1)} = [s_1, s_2, \dots, s_N], \alpha, \beta$
        \STATE Initialize an empty decision diagram
        \STATE Label node $(0,0)$ as $\top$
        \FOR{$i=0$ {\bfseries to} $M$, $j=0$ {\bfseries to} $N$}
            \IF[\textcolor{matchcolor}{~~\# Match $s_j$ and $g_i$}]{$i> 0$ \AND $j > 0$ \AND $g_i = s_j$}
                \STATE Add edge $(i-1,j-1) \to (i,j)$, weight $1$, label $\vb_j := 0$
                \STATE Add edge $(i-1,j-1) \to (i,j)$, weight $\alpha$, label $\vb_j := 1$
            \ENDIF
            \IF[\textcolor{insertcolor}{~~\# Insert $g_i$}]{$i > 0$}
                \STATE Add edge $(i-1,j) \to (i,j)$, weight $0$
            \ENDIF
            \IF[\textcolor{deletecolor}{~~\# Delete $s_j$}]{$j > 0$}
                \STATE Add edge $(i,j-1) \to (i,j)$, weight $-1$, label $\vb_j := 0$
                \STATE Add edge $(i,j-1) \to (i,j)$, weight $-\beta$, label $\vb_j := 1$
            \ENDIF
        \ENDFOR
        \STATE Label node $(M,N)$ as $\bot$
        \RETURN the diagram
        \end{algorithmic}
        \end{minipage}
        \end{algorithm}
        \end{minipage}
        }
\null\hfill
        \scalebox{0.74}{
        \begin{minipage}{0.28\linewidth}
        {
\begin{tikzpicture}[
roundnode/.style={circle, draw=green!60, fill=green!5, very thick, minimum size=7mm},
squarednode/.style={rectangle, draw=red!60, fill=red!5, very thick, minimum size=5mm},
namednode/.style={circle, draw=black!60, fill=black!5, very thick, inner sep=2pt},
anonnode/.style={circle,fill=black,inner sep=2pt},
protolabel/.style={node font=\small,text height=1.5ex, text depth=.25ex},
targetlabel/.style={node font=\small,text height=1.5ex, text depth=.25ex,rotate=-90},
insert/.style={->,thick,draw=insertcolor},
deleteSure/.style={->,thick,draw=deletecolor,bend left=20},
deleteUnsure/.style={->,dashed,thick,draw=deletecolor,bend right=20},
matchSure/.style={->,thick,draw=matchcolor,bend left=10, shorten >= 4pt, shorten <= 4pt},
matchUnsure/.style={->,dashed,thick,draw=matchcolor,bend right=10, shorten >= 4pt, shorten <= 4pt},
deleteSureLabel/.style={above,yshift=3pt,node font=\tiny,text=deletecolor,fill=white,inner sep=1pt},
deleteUnsureLabel/.style={below,yshift=-3pt,node font=\tiny,text=deletecolor,fill=white,inner sep=1pt},
deleteSureUtility/.style={centered,node font=\tiny,text=deletecolor,draw=deletecolor,fill=white,inner sep=1pt,thin,solid},
deleteUnsureUtility/.style={centered,node font=\tiny,text=deletecolor,draw=deletecolor,fill=white,inner sep=1pt,thin,solid},
matchSureLabel/.style={above right,pos=0.5,node font=\tiny,text=matchcolor,fill=white,inner sep=1pt},
matchUnsureLabel/.style={below left,pos=0.5,node font=\tiny,text=matchcolor,fill=white,inner sep=1pt},
matchSureUtility/.style={centered,node font=\tiny,text=matchcolor,draw=matchcolor,fill=white,inner sep=1pt,thin,solid,pos=0.6},
matchUnsureUtility/.style={centered,node font=\tiny,text=matchcolor,draw=matchcolor,fill=white,inner sep=1pt,thin,solid,pos=0.4},
insertUtilityA/.style={node font=\tiny,text=insertcolor,draw=insertcolor,fill=white,inner sep=1pt,thin,solid,pos=0.3},
insertUtilityB/.style={node font=\tiny,text=insertcolor,draw=insertcolor,fill=white,inner sep=1pt,thin,solid,pos=0.7},
insertUtilityN/.style={node font=\tiny,text=insertcolor,draw=insertcolor,fill=white,inner sep=1pt,thin,solid,pos=0.5},
override1matchUnsureLabel/.style={pos=0.4},
override1matchUnsureUtility/.style={pos=0.35},
x=1.2cm,y=1.7cm
]

\node[protolabel] (p0) at (0.5,0.5) {$s_{1}=\text{a}$};
\node[protolabel] (p1) at (1.5,0.5) {$s_{2}=\text{b}$};
\node[protolabel] (p2) at (2.5,0.5) {$s_{3}=\text{c}$};
\node[targetlabel] (t2) at (3.35, -0.5) {$g_{1}=\text{a}$};
\node[targetlabel] (t2) at (3.35, -1.5) {$g_{2}=\text{c}$};
\node[targetlabel] (t2) at (3.35, -2.5) {$g_{3}=\text{b}$};
\node[targetlabel] (t2) at (3.35, -3.5) {$g_{4}=\text{d}$};
\node[namednode] (n0-0) at (0,0) {$\top$};
\node[anonnode] (n0-1) at (0,-1) {};
\node[anonnode] (n0-2) at (0,-2) {};
\node[anonnode] (n0-3) at (0,-3) {};
\node[anonnode] (n0-4) at (0,-4) {};
\node[anonnode] (n1-0) at (1,0) {};
\node[anonnode] (n1-1) at (1,-1) {};
\node[anonnode] (n1-2) at (1,-2) {};
\node[anonnode] (n1-3) at (1,-3) {};
\node[anonnode] (n1-4) at (1,-4) {};
\node[anonnode] (n2-0) at (2,0) {};
\node[anonnode] (n2-1) at (2,-1) {};
\node[anonnode] (n2-2) at (2,-2) {};
\node[anonnode] (n2-3) at (2,-3) {};
\node[anonnode] (n2-4) at (2,-4) {};
\node[anonnode] (n3-0) at (3,0) {};
\node[anonnode] (n3-1) at (3,-1) {};
\node[anonnode] (n3-2) at (3,-2) {};
\node[anonnode] (n3-3) at (3,-3) {};
\node[namednode] (n3-4) at (3,-4) {$\bot$};
\draw[insert] (n0-0) edge node[insertUtilityA]{$0$}  (n0-1);
\draw[insert] (n0-1) edge node[insertUtilityN]{$0$}  (n0-2);
\draw[insert] (n0-2) edge node[insertUtilityN]{$0$}  (n0-3);
\draw[insert] (n0-3) edge node[insertUtilityN]{$0$}  (n0-4);
\draw[insert] (n1-0) edge node[insertUtilityB]{$0$}  (n1-1);
\draw[insert] (n1-1) edge node[insertUtilityN]{$0$}  (n1-2);
\draw[insert] (n1-2) edge node[insertUtilityA]{$0$}  (n1-3);
\draw[insert] (n1-3) edge node[insertUtilityN]{$0$}  (n1-4);
\draw[insert] (n2-0) edge node[insertUtilityN]{$0$}  (n2-1);
\draw[insert] (n2-1) edge node[insertUtilityA]{$0$}  (n2-2);
\draw[insert] (n2-2) edge node[insertUtilityB]{$0$}  (n2-3);
\draw[insert] (n2-3) edge node[insertUtilityN]{$0$}  (n2-4);
\draw[insert] (n3-0) edge node[insertUtilityN]{$0$}  (n3-1);
\draw[insert] (n3-1) edge node[insertUtilityB]{$0$}  (n3-2);
\draw[insert] (n3-2) edge node[insertUtilityN]{$0$}  (n3-3);
\draw[insert] (n3-3) edge node[insertUtilityN]{$0$}  (n3-4);
\draw[matchSure] (n0-0) edge node[matchSureLabel]{$\vb_1{=}0$} node[matchSureUtility]{$1$} (n1-1);
\draw[matchUnsure] (n0-0) edge node[matchUnsureLabel,override1matchUnsureLabel]{$\vb_1{=}1$} node[matchUnsureUtility,override1matchUnsureUtility]{$.7$} (n1-1); %
\draw[matchSure] (n1-2) edge node[matchSureLabel]{$\vb_2{=}0$} node[matchSureUtility]{$1$} (n2-3);
\draw[matchUnsure] (n1-2) edge node[matchUnsureLabel]{$\vb_2{=}1$} node[matchUnsureUtility]{$.7$} (n2-3);
\draw[matchSure] (n2-1) edge node[matchSureLabel]{$\vb_3{=}0$} node[matchSureUtility]{$1$} (n3-2);
\draw[matchUnsure] (n2-1) edge node[matchUnsureLabel]{$\vb_3{=}1$} node[matchUnsureUtility]{$.7$} (n3-2);
\draw[deleteSure] (n0-0) edge node[deleteSureLabel]{$\vb_1{=}0$} node[deleteSureUtility]{$\smminus 1$} (n1-0);
\draw[deleteUnsure] (n0-0) edge node[deleteUnsureLabel]{$\vb_1{=}1$} node[deleteSureUtility]{$\smminus .3$} (n1-0);
\draw[deleteSure] (n0-1) edge node[deleteSureLabel]{$\vb_1{=}0$} node[deleteSureUtility]{$\smminus 1$} (n1-1);
\draw[deleteUnsure] (n0-1) edge node[deleteUnsureLabel]{$\vb_1{=}1$} node[deleteSureUtility]{$\smminus .3$} (n1-1);
\draw[deleteSure] (n0-2) edge node[deleteSureLabel]{$\vb_1{=}0$} node[deleteSureUtility]{$\smminus 1$} (n1-2);
\draw[deleteUnsure] (n0-2) edge node[deleteUnsureLabel]{$\vb_1{=}1$} node[deleteSureUtility]{$\smminus .3$} (n1-2);
\draw[deleteSure] (n0-3) edge node[deleteSureLabel]{$\vb_1{=}0$} node[deleteSureUtility]{$\smminus 1$} (n1-3);
\draw[deleteUnsure] (n0-3) edge node[deleteUnsureLabel]{$\vb_1{=}1$} node[deleteSureUtility]{$\smminus .3$} (n1-3);
\draw[deleteSure] (n0-4) edge node[deleteSureLabel]{$\vb_1{=}0$} node[deleteSureUtility]{$\smminus 1$} (n1-4);
\draw[deleteUnsure] (n0-4) edge node[deleteUnsureLabel]{$\vb_1{=}1$} node[deleteSureUtility]{$\smminus .3$} (n1-4);
\draw[deleteSure] (n1-0) edge node[deleteSureLabel]{$\vb_2{=}0$} node[deleteSureUtility]{$\smminus 1$} (n2-0);
\draw[deleteUnsure] (n1-0) edge node[deleteUnsureLabel]{$\vb_2{=}1$} node[deleteSureUtility]{$\smminus .3$} (n2-0);
\draw[deleteSure] (n1-1) edge node[deleteSureLabel]{$\vb_2{=}0$} node[deleteSureUtility]{$\smminus 1$} (n2-1);
\draw[deleteUnsure] (n1-1) edge node[deleteUnsureLabel]{$\vb_2{=}1$} node[deleteSureUtility]{$\smminus .3$} (n2-1);
\draw[deleteSure] (n1-2) edge node[deleteSureLabel]{$\vb_2{=}0$} node[deleteSureUtility]{$\smminus 1$} (n2-2);
\draw[deleteUnsure] (n1-2) edge node[deleteUnsureLabel]{$\vb_2{=}1$} node[deleteSureUtility]{$\smminus .3$} (n2-2);
\draw[deleteSure] (n1-3) edge node[deleteSureLabel]{$\vb_2{=}0$} node[deleteSureUtility]{$\smminus 1$} (n2-3);
\draw[deleteUnsure] (n1-3) edge node[deleteUnsureLabel]{$\vb_2{=}1$} node[deleteSureUtility]{$\smminus .3$} (n2-3);
\draw[deleteSure] (n1-4) edge node[deleteSureLabel]{$\vb_2{=}0$} node[deleteSureUtility]{$\smminus 1$} (n2-4);
\draw[deleteUnsure] (n1-4) edge node[deleteUnsureLabel]{$\vb_2{=}1$} node[deleteSureUtility]{$\smminus .3$} (n2-4);
\draw[deleteSure] (n2-0) edge node[deleteSureLabel]{$\vb_3{=}0$} node[deleteSureUtility]{$\smminus 1$} (n3-0);
\draw[deleteUnsure] (n2-0) edge node[deleteUnsureLabel]{$\vb_3{=}1$} node[deleteSureUtility]{$\smminus .3$} (n3-0);
\draw[deleteSure] (n2-1) edge node[deleteSureLabel]{$\vb_3{=}0$} node[deleteSureUtility]{$\smminus 1$} (n3-1);
\draw[deleteUnsure] (n2-1) edge node[deleteUnsureLabel]{$\vb_3{=}1$} node[deleteSureUtility]{$\smminus .3$} (n3-1);
\draw[deleteSure] (n2-2) edge node[deleteSureLabel]{$\vb_3{=}0$} node[deleteSureUtility]{$\smminus 1$} (n3-2);
\draw[deleteUnsure] (n2-2) edge node[deleteUnsureLabel]{$\vb_3{=}1$} node[deleteSureUtility]{$\smminus .3$} (n3-2);
\draw[deleteSure] (n2-3) edge node[deleteSureLabel]{$\vb_3{=}0$} node[deleteSureUtility]{$\smminus 1$} (n3-3);
\draw[deleteUnsure] (n2-3) edge node[deleteUnsureLabel]{$\vb_3{=}1$} node[deleteSureUtility]{$\smminus .3$} (n3-3);
\draw[deleteSure] (n2-4) edge node[deleteSureLabel]{$\vb_3{=}0$} node[deleteSureUtility]{$\smminus 1$} (n3-4);
\draw[deleteUnsure] (n2-4) edge node[deleteUnsureLabel]{$\vb_3{=}1$} node[deleteSureUtility]{$\smminus .3$} (n3-4);

\end{tikzpicture}
}
        \end{minipage}
        }

    \caption{
    An example of an uncertainty-aware edit-distance-based utility function, and of the correspondence between dynamic programming and decision diagram construction.
    \cref{alg:simple-edit-distance-dp} shows a dynamic programming implementation of the utility function $u(g,s)$ described in \cref{sec:problem-statement}, taking as input a target goal state $g$, a fixed suggestion $s$, and weights $\alpha$ and $\beta$ for correct and deleted \LowConf{} tokens, respectively.
    We assign positive utility to correctly predicted tokens instead of penalizing inserts so that an empty suggestion has zero utility. Setting $\alpha < 1$ and $\beta < 1$ allows us to approximate the effect of users double-checking \LowConf{} regions: \LowConf{} tokens provide less utility than \HighConf{} tokens if kept but are easier to delete if wrong. 
    \cref{alg:simple-edit-distance-decision-diagram} is an expanded version of this utility function that enables simultaneously searching over the $c_j$ by building a decision diagram, as described in \cref{sec:representing-as-decision-diagrams}. On the right, we show an example decision diagram obtained by running \cref{alg:simple-edit-distance-decision-diagram} with $g^{(1)} = [a, b, c]$, $g^{(k)} = [a, c, b, d]$, $\alpha = 0.7$ and $\beta = 0.3$, colored based on the algorithm steps. Note that our experiments use a more complex utility function, described in \cref{sec:variants} and \cref{app:utility-functions}.
    }
    \label{fig:decision-diagrams}
\end{figure*}
\subsection{Approximating True Intents With Model Samples}\label{sec:approx-true-intents-model-samples}

We start by assuming that we have access to a well-calibrated generative model $\tilde{p}_\theta(G | X)$ that predicts a distribution of plausible goals in a given context. For instance, $\tilde{p}_\theta(G | X)$ could be a language model trained to produce completions of a partial file. Previous work has shown that samples from such a model can give a good proxy for true uncertainty in a generative model as long as the model is well calibrated \citep{eikema2020bayes-risk-decoding,Ott2018AnalyzingUI}.

As such, we can treat the model $\tilde{p}_\theta(G | X)$ as a proxy for the true conditional distribution $p(G|X)$, and try to find
\begin{equation}
\tilde{s}^* = \argmax_{s \in \mathcal{S}} \E_{g \sim \tilde{p}_\theta(G|X = x)}[u(g, s)].
\label{eqn:objective-model-proxy}
\end{equation}
Intuitively, if we find a suggestion $\tilde{s}^*$ that is reliably useful across the high-likelihood goals under $\tilde{p}_\theta(G | X=x)$, and any sample from $p$ should also have high likelihood under $\tilde{p}_\theta$ (e.g. due to training with the cross-entropy objective), we can hope that such a suggestion is also useful for the true user intent (a sample from $p(G|X=x)$).

It is still intractable to exactly find $\tilde{s}^*$, due to the exponentially large set of possible user intents and possible suggestions. We thus search over a restricted space $\mathcal{S}(g^{(1)}) \subset \mathcal{S}$ derived from one of the model outputs $g^{(1)}$ (which we call the \emph{suggestion prototype}), and use a Monte-Carlo estimate over $K$ independent samples $g^{(k)} \sim \tilde{p}_\theta(G | X=x)$ from the model to estimate utility, similar to the minimum-Bayes-risk decoding strategy proposed by  \citet{eikema2020bayes-risk-decoding}:
\begin{equation}
\tilde{s} = \argmax_{s \in \mathcal{S}(g^{(1)})} \frac{1}{K}\sum_{k=1}^K u(g^{(k)}, s),
\label{eqn:objective-finite-sample}
\end{equation}
Here $K$ is a hyperparameter that determines how many samples we optimize over. For the space of confidence-aware suggestions described in \cref{sec:problem-statement}, we can set
\[
\mathcal{S}(g^{(1)}) = \{[(g^{(1)}_1,c_1), \dots, (g^{(1)}_N,c_N)] : c_i \in \cC\},
\]
which corresponds to taking the suggestion tokens $s_i$ from $g^{(1)}$ and  just searching over the confidence markers $c_i$.

\subsection{Decomposing Into Independent Subproblems}\label{subsec:dual-decomposition-decision-diagrams}
A standard technique for optimizing sums over combinatorial discrete spaces such as $\mathcal{S}(g^{(1)})$ is \emph{dual decomposition}. We give a brief overview of dual decomposition as it applies to our problem; see \citet{sontag2011introduction} and \citet{rush2012tutorial} for an in-depth introduction.

We start by choosing an embedding $\phi$ of the search space $\mathcal{S}(g^{(1)})$ into a space of $d$-dimensional binary vectors, such that each possible suggestion $s \in \mathcal{S}(g^{(1)})$ maps to a unique vector $\vb = \phi(s) \in \{0, 1\}^{d}$; here $d$ may depend on $g^{(1)}$ and the chosen embedding $\phi$. 
We then rewrite our utility function $u(g^{(k)}, s)$ as a function $w^{(k)}(\phi(s))$ of those binary vectors $\vb = \phi(s)$, obtaining the equivalent optimization problem
\[
\tilde{\vb} = \argmax_{\vb \in \{0, 1\}^{d}} \sum_{k=1}^K w^{(k)}(\vb),
\]
from which we can recover the solution to \cref{eqn:objective-finite-sample} by inverting the embedding using information from $g^{(1)}$, i.e. by setting $\tilde{s} = \phi^{-1}(\tilde{\vb})$.

Note that the embedding function $\phi$ will usually depend on the set of suggestions being considered. Taking the set $\mathcal{S}(g^{(1)})$ from the previous section as an example, if $g^{(1)}$ has $N$ tokens we might choose $d=N$ and define $\vb = \phi(s)$ by setting $\vb_i = 1$ whenever $c_i = \LowConf{}$ in the suggestion $s = [(g^{(1)}_1,c_1), \dots, (g^{(1)}_N,c_N)] \in \mathcal{S}(g^{(1)})$. (Our experiments use a somewhat more complex embedding function, described in \cref{app:utility-functions}.)

Next, we reinterpret constrained optimization problem over copies of $\vb$, i.e. as
\begin{align}
U
= \max_{\vb^{(1)},\dots,\vb^{(K)} \in \{0, 1\}^{d}} \sum_{k=1}^K w^{(k)}(\vb^{(k)}),
\label{eqn:objective-binary-vector}
\end{align}
subject to $\vb^{(1)} = \vb^{(2)} = \dots = \vb^{(K)}$. Finally, we introduce a set of Lagrange multipliers $\vlambda^{(k)} \in \R^d$ with $\sum_k \vlambda^{(k)}  = 0$ to relax these equality constraints:
\begin{gather}
W^{(k)}(\vb^{(k)}, \vlambda^{(k)}) = w^{(k)}(\vb^{k}) + \vlambda^{(k)} \cdot \vb^{(k)},
\\
L(\vlambda^{(1:K)}) = \sum_{k=1}^K \max_{\vb^{(k)}} W^{(k)}(\vb^{(k)}, \vlambda^{(k)})\ge U.\label{eqn:dual-problem}
\end{gather}
This is known as the Lagrangian dual problem:
$L(\vlambda^{(1:K)})$ is a convex function of $\vlambda$ and an upper bound on $U$, and our goal is to find $\argmin_{\vlambda} L(\vlambda^{(1:K)})$.
If we can find $\vlambda^{(1:K)}$ such that $\vb^{(1)} = \vb^{(2)} = \dots = \vb^{(K)}$ in \cref{eqn:dual-problem}, the bound is tight and we recover the solution to \cref{eqn:objective-binary-vector}. (Note that this bound is not necessarily tight: we may have a nonzero duality gap $\min_\vlambda L(\vlambda) - U >0$.)
The key advantage of this formulation is that the only interaction between terms in the sum is the constraint $\sum_k \vlambda^{(k)}  = 0$, which we can interpret as ``messages'' between subproblems. We can thus alternate between independently optimizing over the $\vb^{(k)}$ for each $W^{(k)}$ term, and adjusting the $\vlambda^{(k)}$ (``message-passing'') to tighten the dual bound by increasing agreement of the $\vb^{(k)}$.

One efficient optimization algorithm of this form is ``max-marginal-averaging'', a version of coordinate descent described by \citet{lange2021efficient}. It works by iterating through variable indices $i$,
computing the \emph{max-marginals}
\begin{align*}
m^{(k)}_{i := \beta} &= \max_{\substack{\vb^{(k)} ~\text{s.t.}~ \vb^{(k)}_i = \beta}} W^{(k)}(\vb^{(k)},\vlambda^{(k)}), ~~~~ \beta \in \{0, 1\}
\end{align*}\\[-1em]
(which measure the utility of fixing $\vb^{(k)}_i$ to $\beta = 0$ or $\beta = 1$),
setting $\delta^{(k)}_{i} = m^{(k)}_{i := 1} - m^{(k)}_{i := 0}$,
and then updating
\begin{align}
\vlambda^{(k)}_i \gets \vlambda^{(k)}_i
&- \delta^{(k)}_{i}
+ \frac{1}{K} \sum_{k'} \delta^{(k')}_{i}\label{eqn:max-marginal-averaging}
\end{align}\\[-1em]
This update ensures $\delta^{(k)}_{i} = \delta^{(k')}_{i}$ for all $k, k'$, which implies that the same choice ($\vb^{(k)}_i := 0$ or $\vb^{(k)}_i := 1$) is optimal for every $k$ and by the same amount. This is a coordinate descent update for $L(\vlambda^{(1:K)})$ with respect to the $\vlambda^{(1:K)}_i$ \citep{lange2021efficient,werner2020relative}, and applying it produces a monotonically decreasing upper bound on $U$.

\subsection{Expanding Utility Functions To Decision Diagrams}\label{sec:representing-as-decision-diagrams}
It remains to show how to efficiently compute the updates in \cref{eqn:max-marginal-averaging} corresponding to our objective in \cref{eqn:objective-finite-sample}. Our key idea is to focus on a family of utility functions that can be computed using an edit-distance-like dynamic program,
and rewrite them in a form that enables us to simultaneously search over edit sequences, which are different for each subproblem, and confidence annotations, which must be chosen consistently across all subproblems.

\cref{fig:decision-diagrams} gives an example of this transformation for the utility function introduced in \cref{sec:problem-statement}.
We start with \cref{alg:simple-edit-distance-dp}, which computes an edit-distance-based utility $u(g,s)$ for a specific suggestion $s$ and a specific vector of confidence annotations $c$ by searching over possible alignments of $s$ and $g$.
We then extend this implementation to \cref{alg:simple-edit-distance-decision-diagram}, which additionally searches over confidence annotations by embedding both the sequence of edits and the sequence of confidence annotations into a single \emph{binary decision diagram} (BDD). Finding the maximum-utility path in this diagram simultaneously computes both the optimal alignment between $s$ and $g$ and the optimal confidence annotations $c_i$, and we can reconstruct the confidence annotations by following the path and setting $c_i = \LowConf{}$ whenever we encounter an edge labeled $\vb_i := 1$ (inverting the original embedding $\phi$).

We can now build a system of decision diagrams by constructing a separate BDD for each model sample $g^{(k)}$, and use this system to solve the optimization problem in \cref{eqn:dual-problem}. Specifically, we can compute the max-marginals $m^{(k)}_{i := \beta}$ for a given variable $\vb_i$ and subproblem $k$ by traversing the diagram for subproblem $k$ and separately considering paths that assign $\vb_i := 0$ and $\vb_i := 1$. The messages $\vlambda^{(k)}_i$ can then be incorporated by modifying the costs of all edges in subproblem $k$ that assign $\vb_i := 1$, which biases that subproblem's search to prefer confidence annotations that are consistent with the choices of other subproblems.

In more detail, we follow \citet{lange2021efficient} and use the BDD representation to run a sequence of max-marginal-averaging updates until $L(\vlambda^{(1:K)})$ stops improving. If all subproblems agree on the optimal assignment $\vb$ (e.g. \cref{eqn:dual-problem} is tight), the maximum-utility paths for each subproblem now correspond to edit sequences for the same suggestion $\tilde{s} = \phi^{-1}(\vb) \in \cS(g^{(1)})$; we can thus reconstruct the solution to \cref{eqn:objective-finite-sample} by combining information from $\vb$ and the suggestion prototype $g^{(1)}$. If the subproblems do not agree, we instead decode an approximate solution to \cref{eqn:objective-binary-vector} by greedily committing to the most promising assignment for each element of $\vb$ and updating max-marginals to be consistent with the fixed assignments, similar to the heuristic described by \citet{kolmogorov2014new}, and then reconstruct a (possibly suboptimal) suggestion from this guess.
We note that additional subtleties are needed to make the BDD representation efficient and to avoid unnecessary recomputation in the dynamic programs; see \cref{app:decision-diagram-definitions}.

\subsection{Extending the Utility Function}\label{sec:variants}
Our method can be applied to any space of suggestions $\mathcal{S}$ and utility function $u(g, s)$ that can be efficiently represented as decision diagrams. 
In this section we briefly summarize a number of extensions to the basic utility function presented in \cref{alg:simple-edit-distance-dp}. These extensions, which we use for our experiments, enable us to adapt the R-U-SURE system to a variety of tasks without modifying the pretrained language model. (See \cref{app:utility-functions} for details.)

\emph{Incorporating tree structure with hierarchical edits.} When data has a natural tree structure (e.g. an abstract syntax tree for a program), we can incorporate this structure into $u(g, s)$ by requiring that edits respect the tree hierarchy. In particular, we implement a recursive utility function under which entire subtrees are either deleted, inserted, or recursively matched with other subtrees at the same depth.

\emph{Constraining locations of \LowConf{} regions.} Similarly, we may have prior knowledge about which tokens are appropriate to mark as $\LowConf{}$; for instance, we may want to ensure that \LowConf{} tokens align with parsed expressions in the syntax tree. We can enforce this by introducing new binary decision variables that track where \LowConf{} regions start and stop, and including a ``constraint BDD'' which ensures they start and stop in semantically-meaningful positions.

\emph{Adding localization and insertion penalties.} Identifying which location to edit may be more difficult than actually performing the edit, and it may be useful to identify locations at which more code must be inserted even if all of the tokens in the suggestion are likely to be kept. To account for this, we can introduce an additional ``localization penalty'' each time an edit starts, independent of the size of the edit. This encourages our method to group edits into semantically meaningful chunks and to identify locations where missing code may need to be inserted, as long as we allow small \LowConf{} regions to be added in spaces between tokens.

\emph{Searching for prefixes.} We may want to extract only a small portion of the model's initial suggestion, stopping once the uncertainty becomes too high. We can account for this by introducing new decision variables that determine whether or not to truncate the suggestion at various points, and modifying $u$ to stop penalizing edits after the truncation point.

\section{Related Work}
\textbf{Decoding by maximizing utility.}
A variety of sampling-based decoding strategies aiming to minimize Bayes risk have been proposed, with many works applying it to neural machine translation
\citep{eikema2020bayes-risk-decoding%
,bhattacharyya2021energy%
,kumar2004minimum%
,ehling2007minimum%
,muller2021understanding%
,eikema2021sampling%
,freitag2022high%
} and to code generation \citep{li2022alphacode,shi2022natural}.
These approaches generally use a utility function to select one sample from a larger generated set. \citet{gonzalez2015minimum} also explore combining parts of multiple samples to construct a single combined sample, and \citet{Lin2010ARM} propose using Bayes risk for extractive summarization.
Although not framed as utility maximization, self-consistency decoding \citep{Wang2022SelfConsistencyIC,Huang2022LargeLM} also uses samples to identify the most likely correct answer under a model's distribution, and has been shown to improve reasoning ability.

Reinforcement-learning and sequential-decision-making techniques can also be used to maximize conditional expected utility, by interpreting tokens as actions and the utility as a reward
\citep{lampouras2016imitation%
,sun2017deeply%
,chen2018stable%
,prabhavalkar2018minimum-WER%
,keneshloo2019deep%
,leblond2021machine%
}. Many works maximize quality metrics such as BLEU or error-rate, although others have used measures of program correctness
\citep{le2022coderl}
or learned reward models
\citep{ziegler2019fine%
,ouyang2022training%
,bai2022training%
}.
Others have trained models to imitate a more expensive reward-driven search process
\citep{Kuncoro2016DistillingAE%
,Liu2018DistillingKF%
,sabour2019optimal-completion-distillation%
}.

\textbf{Selective and multi-choice prediction.}
One approach to avoid incorrect predictions under uncertainty is \emph{selective classification}, i.e. abstaining from some predictions to minimize overall risk
\citep{chow1957optimum%
,el2010foundations%
,geifman2017selective%
,dong2018confidence%
,ziyin2019deep%
}.
Another approach is to output multiple predictions, e.g.  all classifications with confidence above a threshold
\citep{vovk2005algorithmic,angelopoulos2021gentle},
or an ensemble of structured outputs which approximately covers the true output
\citep{guzman2012multiple%
,guzman2014efficiently%
,prasad2014submodular%
,lee2016stochastic%
,bhattacharyya2018accurate%
,firman2018diversenet%
,Premachandran2014EmpiricalMB%
}.
When the space of possible outputs is very large, uncertain predictions can be compressed by representing multiple sequences as a lattice
\citep{su2017lattice,Sperber2017NeuralLM};
lattice representations have also been used within a Bayes risk framework
\citep{Tromble2008LatticeMB,Xu2010AnIC}.

\textbf{Generating and identifying partial programs.}
A number of works have considered identifying common patterns in source code
\citep{Lozano2010MiningSC%
,Allamanis2014MiningIF%
,Shin2019ProgramSA%
,Sivaraman2021MiningII%
},
as well as generating programs with holes to aid in program synthesis
\citep{Nye2019LearningSketches,Ellis2020DreamCoderGG}.
Most relevant to our current work, \citet{Guo2021Grammformer} propose \textsc{Grammformer}, a generative model for code that produces holes in parts of the syntax tree that are difficult to predict. \textsc{Grammformer} generates code top-down by iteratively expanding nonterminal nodes of a syntax tree, and is trained via a combination of random-order pretraining and RL finetuning, using a regular-expression-based objective that trades off between size and accuracy. In contrast, our approach requires only sample access to a pretrained generative model, can adapt to different utility functions and suggestion types without retraining the model, and can identify regions of uncertainty in both syntax trees and unstructured sequences (e.g. docstrings) without aligning them to a syntax tree derivation.

\textbf{Uncertainty quantification and summarization.}
Past works have compared model-generated sequences to ground truth
\citep{Ott2018AnalyzingUI,Holtzman2019TheCC},
studied the calibration of deep models in general
\citep{Carrell2022CalibrnGeneralizn},
and proposed new mechanisms for training better-calibrated models
\citep{Tran2022PlexTR,Xiao2022UncertaintyQW}.
\citet{Kadavath2022LanguageM} find that some large language models can sometimes answer natural-language questions about the accuracy of their own generated outputs, improving when multiple sampled outputs are included in the prompt, and
\citet{Lin2022TeachingMT} show that language models can be fine-tuned to output calibrated uncertainty estimates.
Our work relies on the calibration and sample-quality of the base intent model, but focuses on exposing this uncertainty to end-users.
Also related are works which use attention and saliency maps to inform users about model behavior
\citep{Stevens2020AnIO, Tenney2020TheLI},
as well as works that visualize per-token probabilities to summarize model uncertainty
\citep{Strobelt2021LMdiffAV,Weisz2021PerfectionNR,Sun2022InvestigatingEO}.
Most relevant to our work, \citet{Vasconcelos2022GenerationPA} found that visualizations of predicted locations of edits are strongly preferred by users over individual token probabilities and also significantly reduce editing time. They used a separate model trained to predict edits for a specific coding problem based on user editing traces; in contrast, our technique can be used to produce annotations without requiring additional model training or edit supervision.

\textbf{Combinatorial optimization.}
Dual decomposition and block coordinate descent/ascent solvers have been applied to a variety of optimization problems, including combinatorial search
\citep{Swoboda2016ADA},
MAP inference \citep{sontag2011introduction}, and NLP tasks \citep{Rush2010OnDD}. Relevant to our work, \citet{Paul2012ImplicitlyIW} and \citet{Peng2015DualDI} use dual decomposition with n-gram features to combine WFSAs, with applications to minimizing Bayes risk.
There has also been recent interest in using binary decision diagrams as representations for combinatorial optimization  \citep{Castro2022DecisionDF}. Our work expands on that of \citet{lange2021efficient} by applying dual decomposition to a larger class of decision diagrams;
see \cref{app:decision-diagram-definitions}.

\section{Experiments}

We evaluate our approach by applying it to three developer assistance tasks, each of which is visualized in \cref{fig:frontpage}. For all tasks, we generate suggestion prototypes and hypothetical intents using a 5B-parameter decoder-only LM trained on 105B tokens of permissively-licensed open-source code from GitHub, and parse them into trees using an error-tolerant bracket-matching pseudo-parser (described in \cref{app:pseudo-parser}). We compare our approach to a number of task-specific baselines, all of which build suggestions $s \in \cS(g^{(1)})$ from the same suggestion prototype, and evaluate how well each method can predict the changes necessary to obtain the final code state from the dataset, measured by our utility function as well as token accuracy.

\subsection{Localizing edits in code suggestions}
\label{sec:experiments:localizing}

\begin{table}[t]
    \centering
{\footnotesize
\begin{tabular}{ccccc@{}c@{}c@{}}
\toprule
{}  & \thead{Utility \\ {\scriptsize (relative)}} & \thead{Est. Util. \\ {\scriptsize (relative)}} & \thead{LOO Util.\\ {\scriptsize (relative)}} & \thead{$F_1$ score\\ {\scriptsize (for \LowConf{})}} \\
\midrule
\scriptsize{\textsc{All Sure}}
&
$\equiv 0$ &
38.00 &
30.35 &
-
\\
\scriptsize{\textsc{Max Unsure}}        &             81.83 &              106.08 &             101.90 &                             12.74 \\
\scriptsize{\textsc{Token Pr. 0.5}}  &             50.83 &               82.63 &              77.10 & 63.03  \\
\scriptsize{\textsc{Token Pr. 0.7}}  &             58.42 &               88.89 &              83.68 & 64.38 \\
\scriptsize{\textsc{Token Pr. 0.9}}  &             66.99 &               95.64 &              90.79 & 61.44 \\
\scriptsize{\textsc{Prefix Pr. 0.5}} &             83.33 &              108.52 &             104.29 & 41.92 \\
\scriptsize{\textsc{Prefix Pr. 0.7}} &             83.45 &              108.27 &             104.05 & 37.26 \\
\scriptsize{\textsc{Prefix Pr. 0.9}} &             83.08 &              107.61 &             103.41 & 29.53\\
\scriptsize{\textsc{Ours} (Region)}         &             \textbf{84.42} &              \textbf{113.82} &             \textbf{109.12} & \textbf{72.14} \\
\bottomrule
\end{tabular}
}

\caption{Breakdown of edit-localization performance.
Methods that perform well on model samples (Est. and LOO Utility) also perform well on the ground truth user intent (Utility); we measure utility relative to labeling the entire suggestion \HighConf{}.
Our approach achieves higher utility and also stronger token level $F_1$ score when interpreting \LowConf{} tokens as predicted edits. 
    }
    \label{tab:uncertainty-regions}
\end{table}

\newcommand\cpp{\textsc{C++}}
\newcommand\java{\textsc{Java}}
\newcommand\javascript{\textsc{Javascript}}
\newcommand\python{\textsc{Python}}
Our first task uses \RUSURE{} to insert confidence annotations around parts of code completion suggestions that users are likely to edit. As discussed in \cref{sec:problem-statement}, we configure our utility function so that \LowConf{} tokens have lower utility if matched but lower penalties if deleted. We additionally enforce hierarchical edits and syntactically-valid \LowConf{} regions and add extra localization penalties using the extensions described in \cref{sec:variants}; see \cref{app:experiment-details-utility-configuration} for the specific configuration we use.

To evaluate our approach, we assemble a collection of 5000 held-out code files for each of the languages \java , \javascript, \cpp\ and \python, and split them into (synthetic) completion contexts $c$ and ground truth intents $g$ using three strategies. One such scheme is \python\ specific, so we obtain 45000 examples in total (see \cref{app:experiment-details} for details). For each example, we sample 31 completions from the language model, then select the sample with the highest likelihood as the suggestion prototype, and use R-U-SURE to mark parts of the parsed tree as \LowConf.
We compare our approach to heuristics based on token probabilities, which insert \LowConf{} regions around tokens whose conditional probability (\textsc{Token Prob}) or total prefix probability (\textsc{Prefix Prob}) is below a threshold; we also try marking everything \HighConf{}, and marking the maximum amount as  \LowConf{} in our syntax-constrained space $\mathcal{S}(g^{(1)})$. (We find that annotating based on token probability can miss high-likelihood tokens that depend on earlier low-likelihood tokens, as shown in \cref{fig:app:example4,fig:app:example6} of \cref{app:example-outputs}.)

\begin{figure}[t]
    \includegraphics[width=0.49\linewidth]{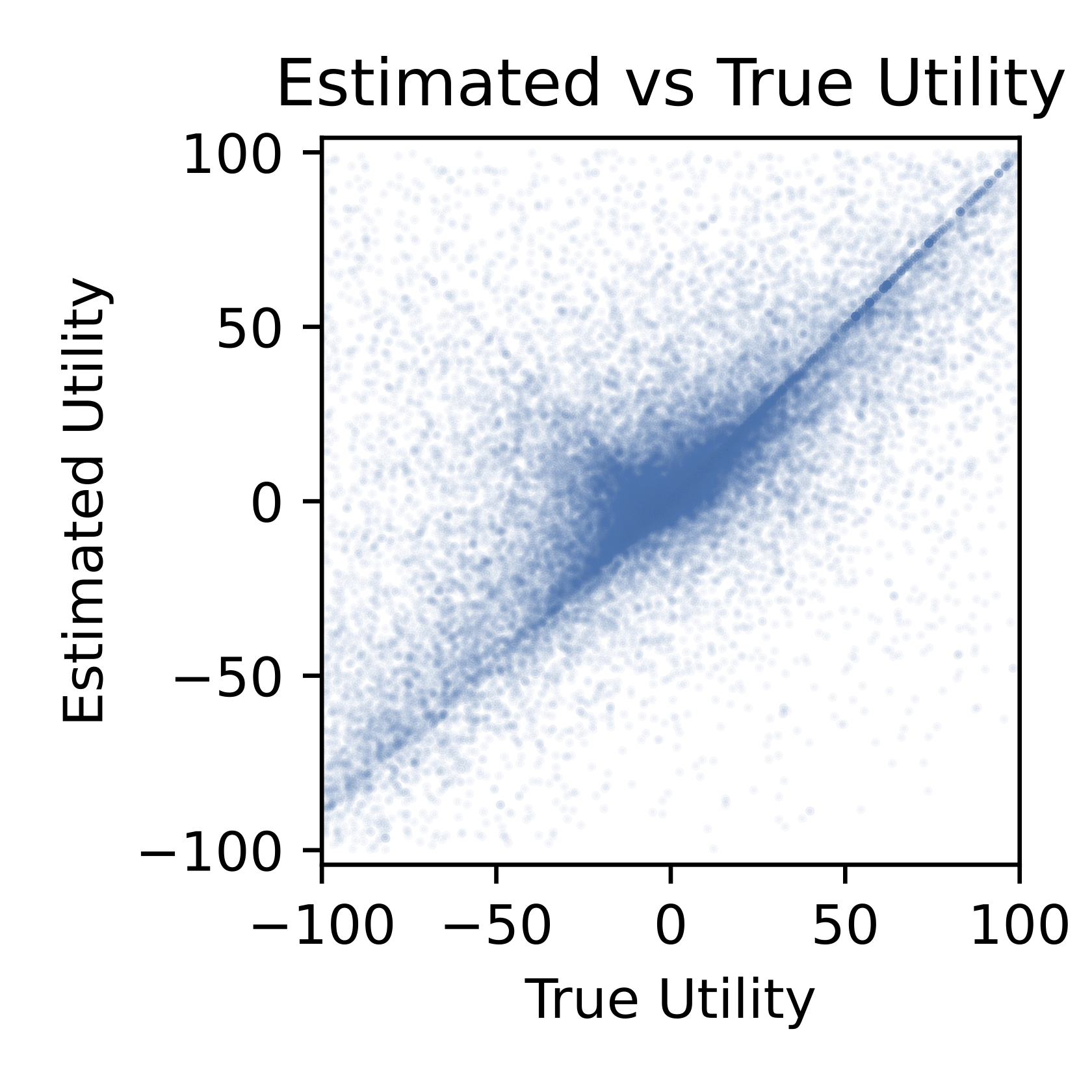}%
    \null\hfill%
    \includegraphics[width=0.49\linewidth,page=4]{figures/results/Regions_5000_32_samples_logits_metrics_UncertaintyRegionsWrapper_144_system_vs_gt.pdf}
    \hfill
    \caption{
    Comparison of true and estimated utility for edit localization task, as a scatterplot and as a rank-histogram \citep{Candille2005EvaluationOP}. The true utility of our approach's suggestions is correlated with the estimate from the model samples, and appears fairly often at any rank within those samples, indicating that the samples are often a good proxy for the true uncertainty in the user's intent. However, some examples have a lower ground-truth utility than estimated, potentially indicative of model miscalibration.
    }
    \label{fig:true-vs-estimated-utility}
\end{figure}

\begin{figure}[t]
  \begin{center}
     \begin{subfigure}[b]{0.22\textwidth}
         \centering
         \includegraphics[width=33mm]{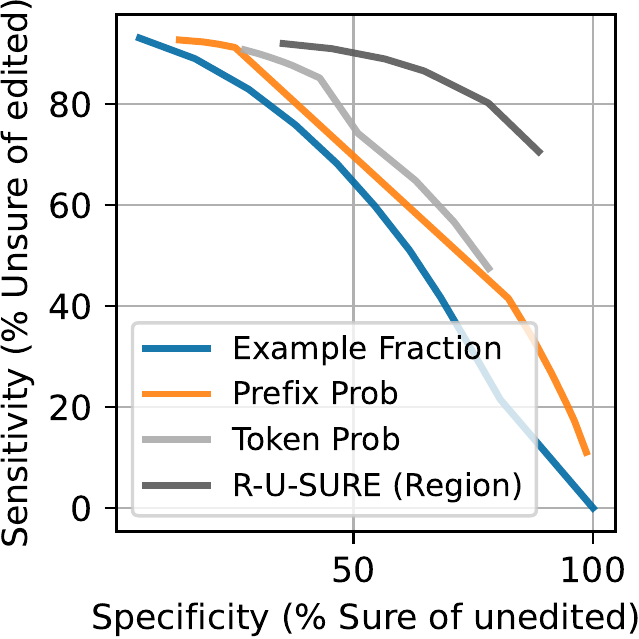}
         \caption{Sensitivity \textit{vs} Specificity.}\label{fig:pareto}
     \end{subfigure} 
     \hfill 
     \begin{subfigure}[b]{0.22\textwidth}
         \centering
         \includegraphics[width=33.5mm,page=1]{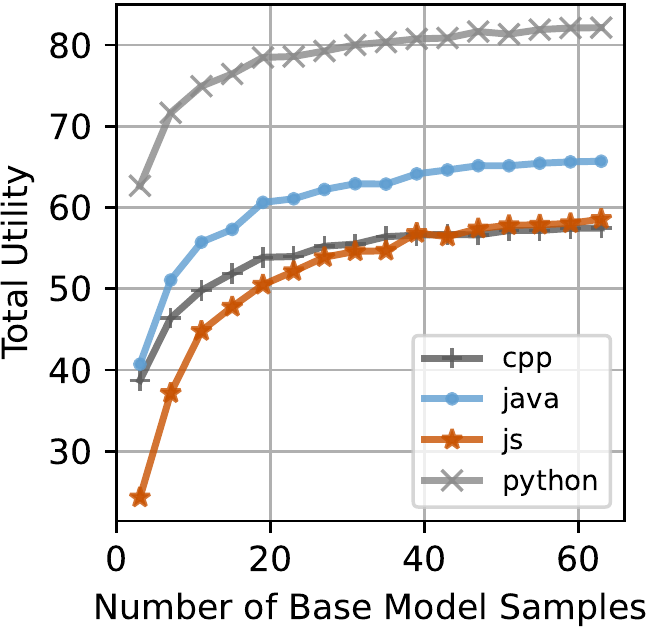}
         \caption{Utility \textit{vs.} sample size $K$.}\label{fig:num-samples}
     \end{subfigure}  
  \end{center}
  \vspace{-1em}
  \caption{Analysis of results for edit localization. \textit{(a)} Token level sensitivity / specificity trade-off across methods; our approach Pareto-dominates the others. (Here ``Example Fraction'' marks a fixed fraction as \LowConf{}.)
 \textit{(b)} Mean utility for our approach w.r.t the ground-truth user intent as a function of the number of samples used in \cref{eqn:objective-finite-sample}, split by programming language.
  }
\end{figure}

We first investigate how well our optimizer of the sample-based approximate objective in \cref{eqn:objective-finite-sample} succeeds at producing a good suggestion for the true unobserved intent $g$, as measured by our utility metric $u$. To this end, in  \cref{tab:uncertainty-regions} we report
the utility $u(g,s)$ of each method's annotated suggestion $s$ based on the true file contents $g$ (Utility), and also our estimate of utility $\frac{1}{K} \sum_{k=1}^K u(g^{(k)},s)$ computed across $K=31$ samples from the model (Est. Util.).
We additionally report the estimated utility $u(g^{(K+1)},s)$ for a held-out model sample $g^{(K+1)}$ which was not used for optimization, denoting this ``Leave-One-Out Utility'' (LOO Util.). We find that methods with high average utility on model samples also achieve high average utility against the true file contents, with our method successfully obtaining high utility in both settings. A more detailed comparison in \cref{fig:true-vs-estimated-utility} reveals that utility on the model samples $g^{(k)}$ and utility on the unobserved intent $g$ are highly correlated for our suggestions, and \cref{fig:num-samples} shows that utility also improves as we optimize over more samples.

Note that our utility metric is only an approximation of the quality of an uncertainty-annotated suggestion, so having higher utility does not necessarily imply that our method produces more useful suggestions. To give more insight and evaluate how well maximizing our utility function truly summarizes the uncertainty of the model, we reinterpret uncertainty region annotations as a binary classification problem, with \LowConf{} tokens being predictions of where users will edit. 
We then compute the sensitivity (fraction of edited code correctly marked \LowConf{}) and specificity (fraction of unedited code marked  \HighConf{}) for all methods with respect to the ground truth $g$. We visualize how these metrics vary as we sweep over the per-token utilities and costs of \LowConf{} tokens, obtaining the Pareto curve in \cref{fig:pareto}; we also summarize the accuracy with $F_1$ scores in \cref{tab:uncertainty-regions}. We find that our approach is better at identifying locations of edits than the baselines, indicating that maximizing our utility metric does produce meaningful uncertainty estimates.

\subsection{Selecting Suggestion Lengths}
One common use of ML model suggestions for both code and natural language applications is to show inline grey ``ghost text'' suggestions as users type in the editor, and allow users to quickly accept the suggestion by pressing a key (often \texttt{tab}) \citep{svyatkovskiy2020intellicode,Barke2022GroundedCH}. In this case, showing longer correct suggestions can accelerate developer productivity, but long incorrect suggestions can slow developers down \citep{Barke2022GroundedCH}.

To apply our approach to this setting, we disable insertion of \LowConf{} annotations, and instead search over prefixes of the prototype suggestion using the truncation variables described in \cref{sec:variants}; we continue to enforce hierarchical edits as well (see \cref{app:experiment-details-utility-configuration}).
We compare our approach to a number of heuristics: predicting a fixed number of lines, predicting until we reach a low-probability token (Token Prob.) or until the total probability is below a threshold (Prefix Prob.), using the heuristic described by \citet{svyatkovskiy2020intellicode} (IntelliCode Compose), and choosing the prefix that maximizes the ratio between the log-probability of the sequence and its length (Max Avg. Log Prob) inspired by the similar heuristics described by \citet{chen2021evaluating} and \citet{zhang2022coder}.

\begin{figure}
    \centering
    \includegraphics[width=0.95\linewidth,trim=0 0.25cm 0 0cm,clip]{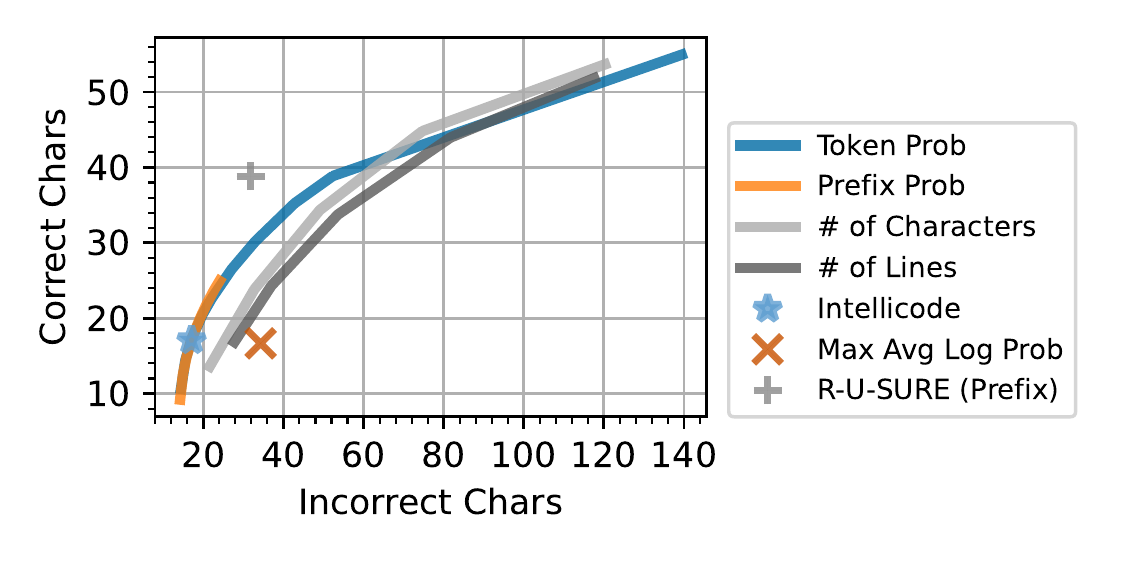}
    \caption{
    Character-level accuracy tradeoff for the suggestion-length task. R-U-SURE obtains a favourable trade-off. (Note that for simplicity we did not vary the hyperparameters of R-U-SURE for this task.)
    }
    \label{fig:prefixpareto}
\end{figure}

\cref{fig:prefixpareto} visualizes the tradeoff between correct and incorrect characters predicted by each method, and shows that our approach achieves a better tradeoff than other prefix-selection baselines.
\cref{tab:appendix:prefix-length} in \cref{app:detailed-experimental-results} shows that our approach also achieves strong results on our utility metric, similar to the edit localization task.

\subsection{API Discovery}
Even if there is not enough information to provide a useful completion suggestion at a specific location, it may still be possible to extract useful information from a generative model's suggestions. As an example of this, we use our method to identify \emph{sequences of function calls} that the user is likely to write, even if the control flow structures around these calls are not predictable; this could be used to preemptively show documentation or type signatures.

We adapt our approach to this setting by extracting a sequence of function and method calls from the model's output, choosing $\mathcal{S}(g_1)$ so that it selects a subset of these calls as \HighConf{}, and defining $u(g,s)$ to find the longest common subsequence between the desired calls in $g$ and the selected calls. Since we expect such suggestions to be used as an auxiliary aid rather than an inline suggestion, we set the utility of correct predictions to be higher than the penalty for incorrect ones (e.g. selected calls in $s$ that were not used in $g$) and give a bonus for predicting tokens not seen before; we also assign zero utility to unselected calls. We implement this by applying the hierarchical edits and region location constraints described in \cref{sec:variants} to a serialized representation of the extracted calls; see \cref{app:experiment-details-utility-configuration} for the specific configuration we use.

We compare our approach against baselines which use all calls in the file or which only predict calls that use identifiers that are not in the context. Results are shown in \cref{tab:api-discovery}; we again find that our approach both achieves strong performance on our utility metric and gives a favorable tradeoff between correct and incorrect predictions.
\begin{table}[t!]
    \centering
    {\footnotesize
    \begin{tabular}{ccccc}
        \toprule
         & \thead{Utility}
         & \thead{Correct\\tokens}
         & \thead{Incorrect\\tokens} \\
        \midrule
\scriptsize{\textsc{Novel Calls}}
&-1.39 &1.18 &3.04 \\
\scriptsize{\textsc{All Calls}}  
&-1.39 &1.18 &3.04 \\
\scriptsize{\textsc{All Calls + LHS + Args}}     
&-8.75 &2.02 &9.64 \\
\scriptsize{\textsc{Ours (Useful Calls)}}
&\textbf{5.10} &3.56 &2.10 \\
\bottomrule
    \end{tabular}
    }
    \caption{
    Comparison of utility and token-level accuracy statistics for API call sequence task. Our approach achieves higher utility by selecting a subset of method calls that is likely to appear, and including the LHS and arguments only when they are predictable.
    }\label{tab:api-discovery}
    \vspace{-1em}%
\end{table}

\section{Discussion}
We have demonstrated that R-U-SURE can flexibly incorporate uncertainty annotations into model suggestions across a variety of developer-assistance tasks, and that these annotations lead to both improved performance on our estimates of utility and also accurate predictions of the locations of edits. Importantly, our approach does not require retraining or fine-tuning the base generative language model, since it decouples the action (showing a suggestion) from the generative prediction task (predicting the user's intent).

A limitation of our approach is that it is restricted to utility functions that can be efficiently decomposed into decision diagrams. This is a good fit for edit-distance-based utility functions, and we believe the same principles could be extended to support multiple confidence levels or suggested alternatives. However, more general types of utility function (e.g. behavioral equivalence) may be difficult to approximate with our technique. We also assume that the base generative model is well-calibrated, and that a modest number of samples from it can summarize the possible edits required. It would be interesting to study how our system behaves with less-calibrated models, and how this changes as the capacity of the base model grows.

Our current implementation of R-U-SURE runs on the CPU using Numba \citep{Lam2015NumbaAL}, and takes between 1 second and 1 minute depending on the number, length, and complexity of the sampled programs (see \cref{app:examples-runtime} for details). Notably, this runtime is dominated by the time to build the decision diagrams rather than the time to run the coordinate-ascent optimizer, likely because our utility function is very general and was designed for flexibility rather than decision-diagram construction efficiency.
Although out of scope of this paper, we have explored distilling the outputs of R-U-SURE into a learned model similar to \citet{Kuncoro2016DistillingAE} and \citet{Kadavath2022LanguageM}, which can then be queried in real time with comparable accuracy to the original R-U-SURE system. Runtime could also be improved by rewriting in a lower-level langauge, specializing the utility function to a specific task, or using GPU acceleration \citep{Abbas2021FastDOGFD}.

More broadly, we are excited by the potential to incorporate user interaction into minimum-Bayes-risk objectives to mitigate harms of model hallucinations. We see our work as a step toward ML-powered assistants that empower users and give appropriately conservative predictions in the presence of uncertainty about user intent and the world at large.

\section*{Acknowledgements}
We would like to thank
Jacob Hegna, Hassan Abolhassani, Jacob Austin, and Marc Rasi for contributing ideas toward early designs of the R-U-SURE system, 
Maxim Tabachnyk, Chris Gorgolewski, Vladimir Pchelin, Yurun Chen, Ilia Krets, Savinee Dancs, Alberto Elizondo, Iris Chu, Ambar Murillo, Ryan McGarry, Paige Bailey, and Kathy Nix for useful discussions and for collaborating on code completion applications of R-U-SURE,
and Miltiadis Allamanis for providing valuable feedback on the paper draft.
We would also like to thank Abhishek Rao, Alex Polozov, Joshua Howland, Kefan Xiao, and Vedant Misra for providing the language models and evaluation data used for our experimental results,
and the members of Google Brain's Machine Learning for Code team for useful feedback
throughout the project.

\bibliography{references}
\bibliographystyle{icml2023}

\appendix
\onecolumn

\clearpage
\newpage
\section{Example Outputs of R-U-SURE}\label{app:example-outputs}
In this section we give examples of the outputs produced by R-U-SURE.

(For these examples, we use an 8B-parameter decoder-only LM, a slightly larger model than used for the main set of experiments, also trained on permissively-licensed open-source code from GitHub. We emphasize that our approach is model-agnostic and can be combined with any generative model.)

\begin{figure}[H]
    \centering
    \includegraphics[page=4,trim=0.75in 2in 0.5in 0.9in,clip,width=0.9\linewidth]{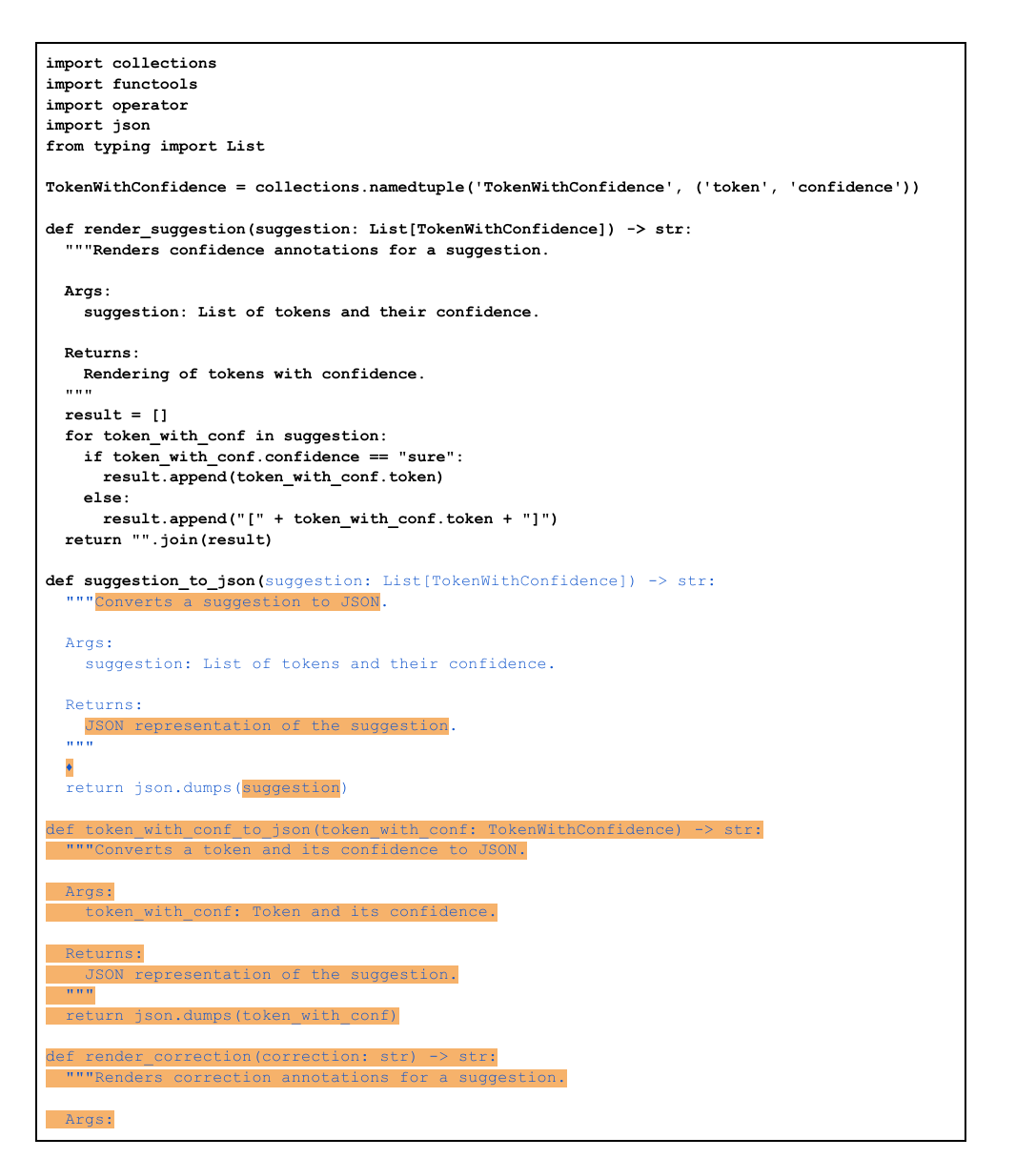}
    \caption{Full prompt and output for the example at the top of \cref{fig:frontpage}. Note that the model has identified the docstring style from the context, and our system can identify which of the words in the docstring are boilerplate. Docstrings are represented as sequences of words and combined using our edit distance utility function. The `$\blacklozenge$' character denotes a location where additional statements might be inserted.
    }
    \label{fig:app:example1}
\end{figure}

\begin{figure}[p]
    \centering
    \includegraphics[page=11,trim=0.9in 2.7in 0.9in 1in,clip,width=\linewidth]{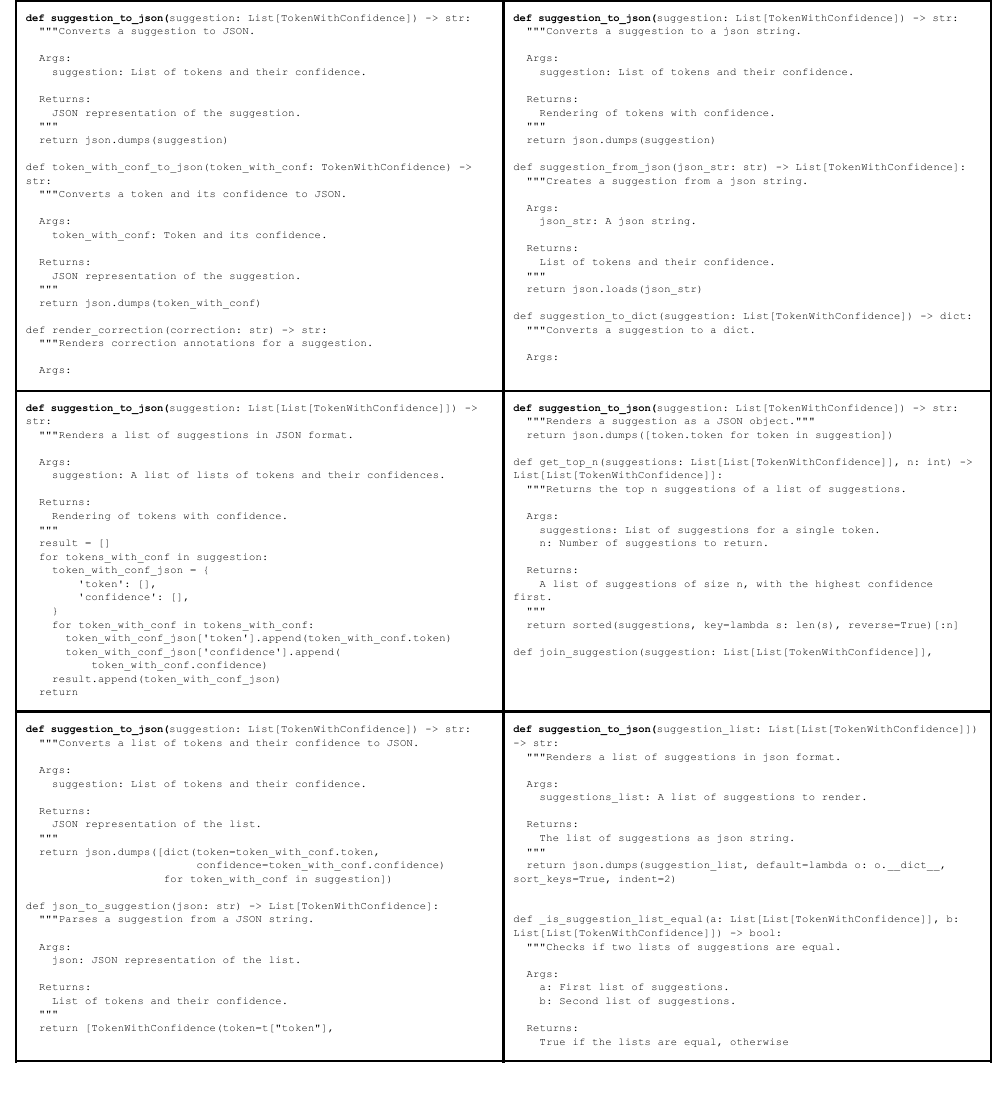}
    \caption{Six of the hypothetical user intents $g^{(1)}, g^{(2)}, \dots, g^{(6)}$ for the example at the top of \cref{fig:frontpage}, generated by sampling from the pretrained model. Full context omitted; see \cref{fig:app:example1}.
    }
    \label{fig:app:example1hypotheticals}
\end{figure}

\begin{figure}[p]
    \centering
    \includegraphics[page=12,trim=0.9in 1.6in 0.9in 1in,clip,width=\linewidth]{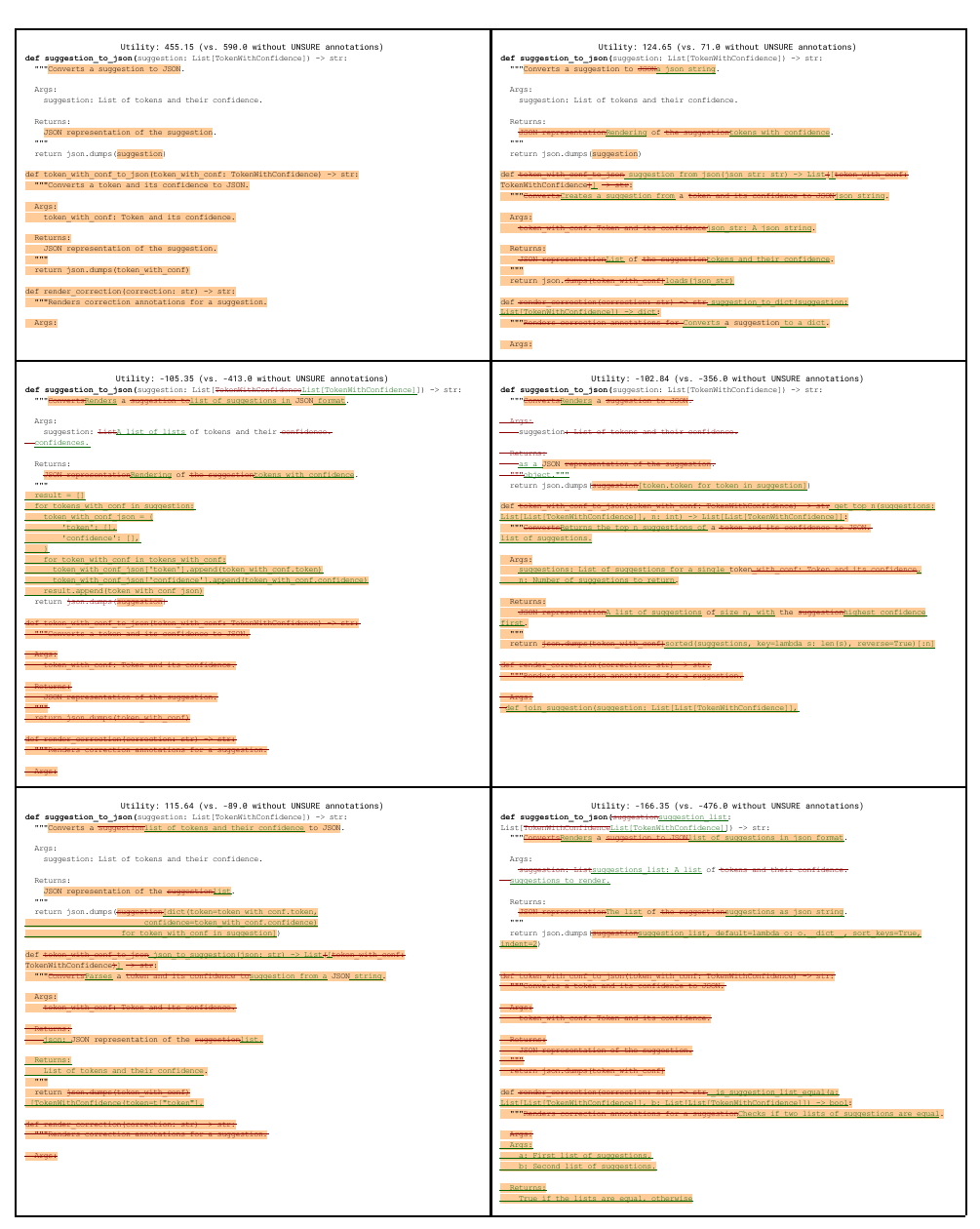}
    \caption{Inferred edits from the output suggestion in \cref{fig:app:example1} to each of the hypothetical user intents in \cref{fig:app:example1hypotheticals}, along with the utility estimates for each when we either insert \LowConf{} regions as shown or require all tokens to be marked \HighConf{}.
    Constant utility shifts do not affect relative utility of different suggestions, so for our results in \cref{tab:uncertainty-regions} and \cref{fig:num-samples}, we report utility relative to marking all tokens as \HighConf{} (i.e. the difference between the two values shown here).
    Note that utility improves when adding \LowConf{} regions for all samples except the first, which was the sample used as the suggestion prototype.
    }
    \label{fig:app:example1edits}
\end{figure}

\begin{figure}[p]
    \centering
    \includegraphics[page=5,trim=0.5in 4.5in 0.5in 0.5in,clip,width=\linewidth]{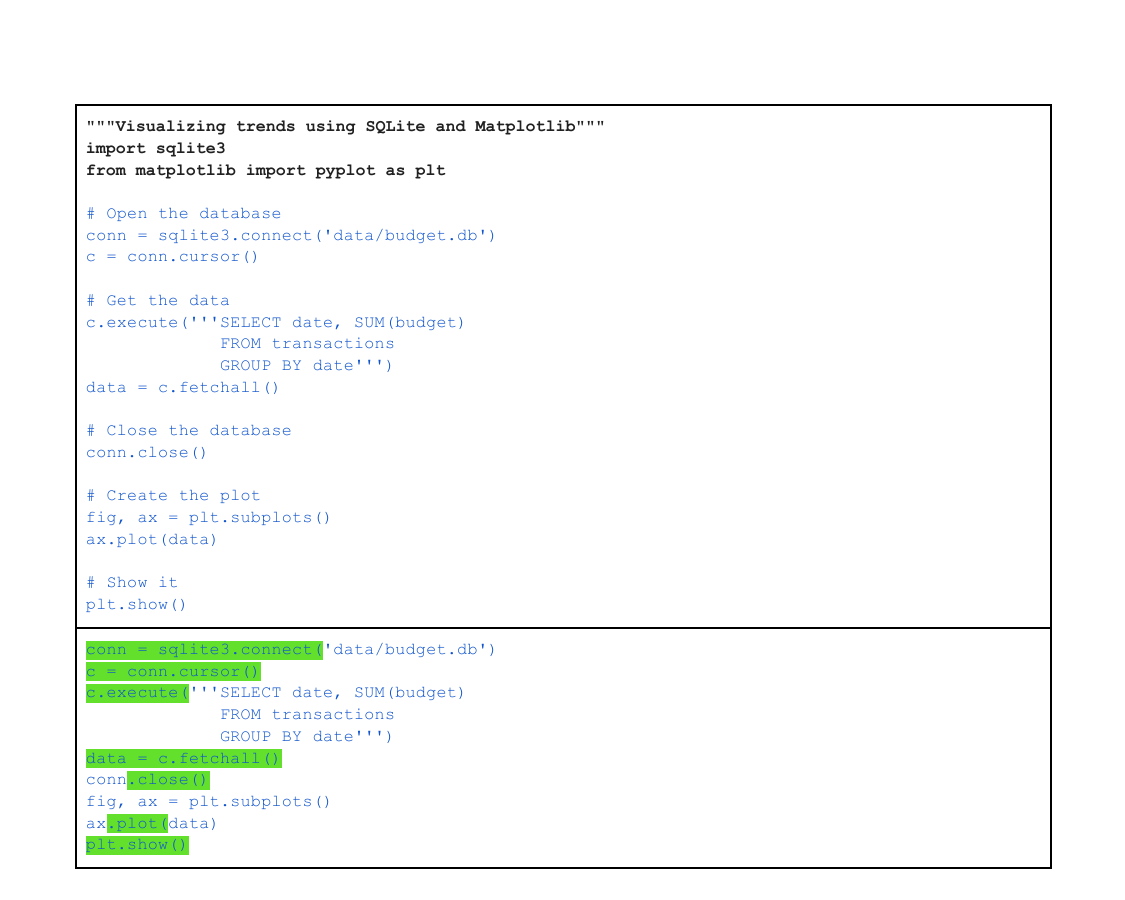}
    \caption{Full prompt and output for the example at the right of \cref{fig:frontpage}. Above, the full generated output of the model. Below, the possible calls we extracted by postprocessing the raw output, with highlighting denoting the calls selected by R-U-SURE. (Note that for this task R-U-SURE operates on this reduced set of calls only.)}
    \label{fig:app:example2}
\end{figure}

\begin{figure}[p]
    \centering
    \includegraphics[page=6,trim=0.5in 5in 0.5in 0.5in,clip,width=\linewidth]{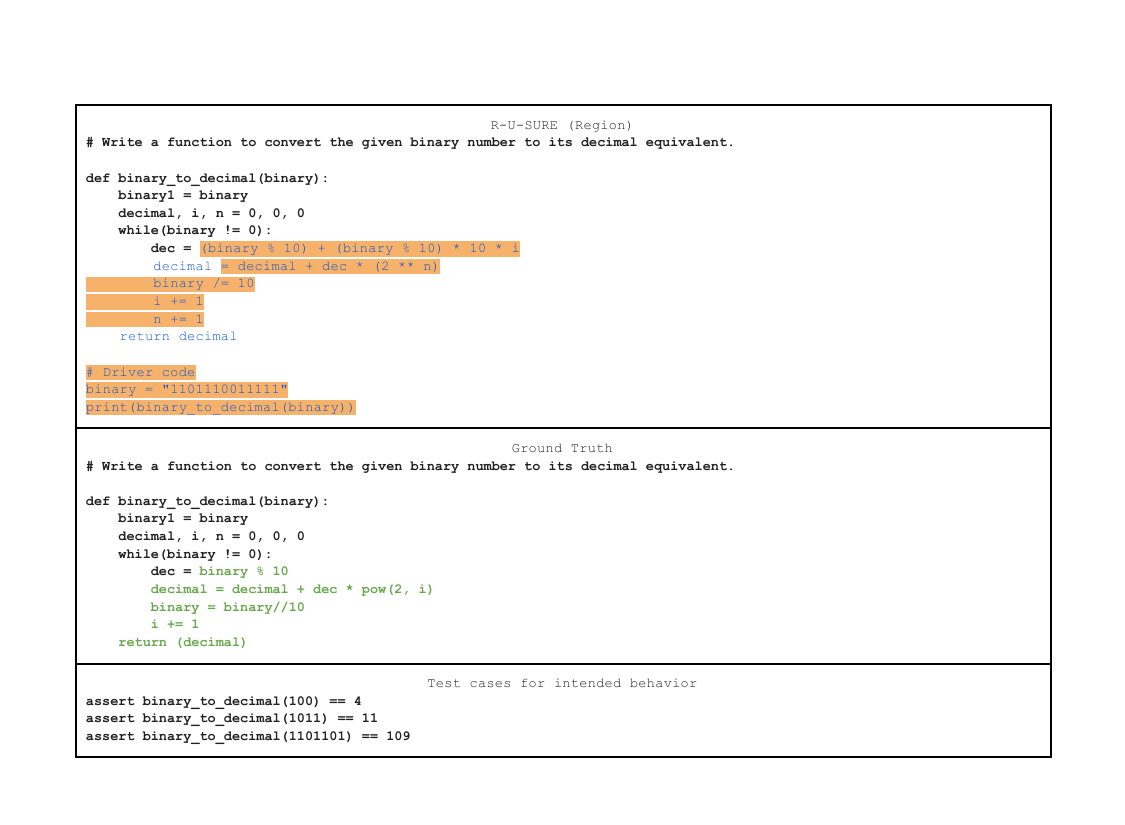}
    \caption{Output of R-U-SURE compared to the ground truth for an example in the Mostly Basic Python Problems benchmark dataset \citep{Austin2021ProgramSW}. We manually selected a location in the MBPP reference solution, then fed the prefix to the model. The model's implementation does not exactly match the intended behavior, but all incorrect parts are highlighted.
    (Note: MBPP examples were not used in our main experimental results.)}
    \label{fig:app:example3}
\end{figure}

\begin{figure}[p]
    \centering
    \includegraphics[page=7,trim=0.5in 3in 0.5in 0.5in,clip,width=\linewidth]{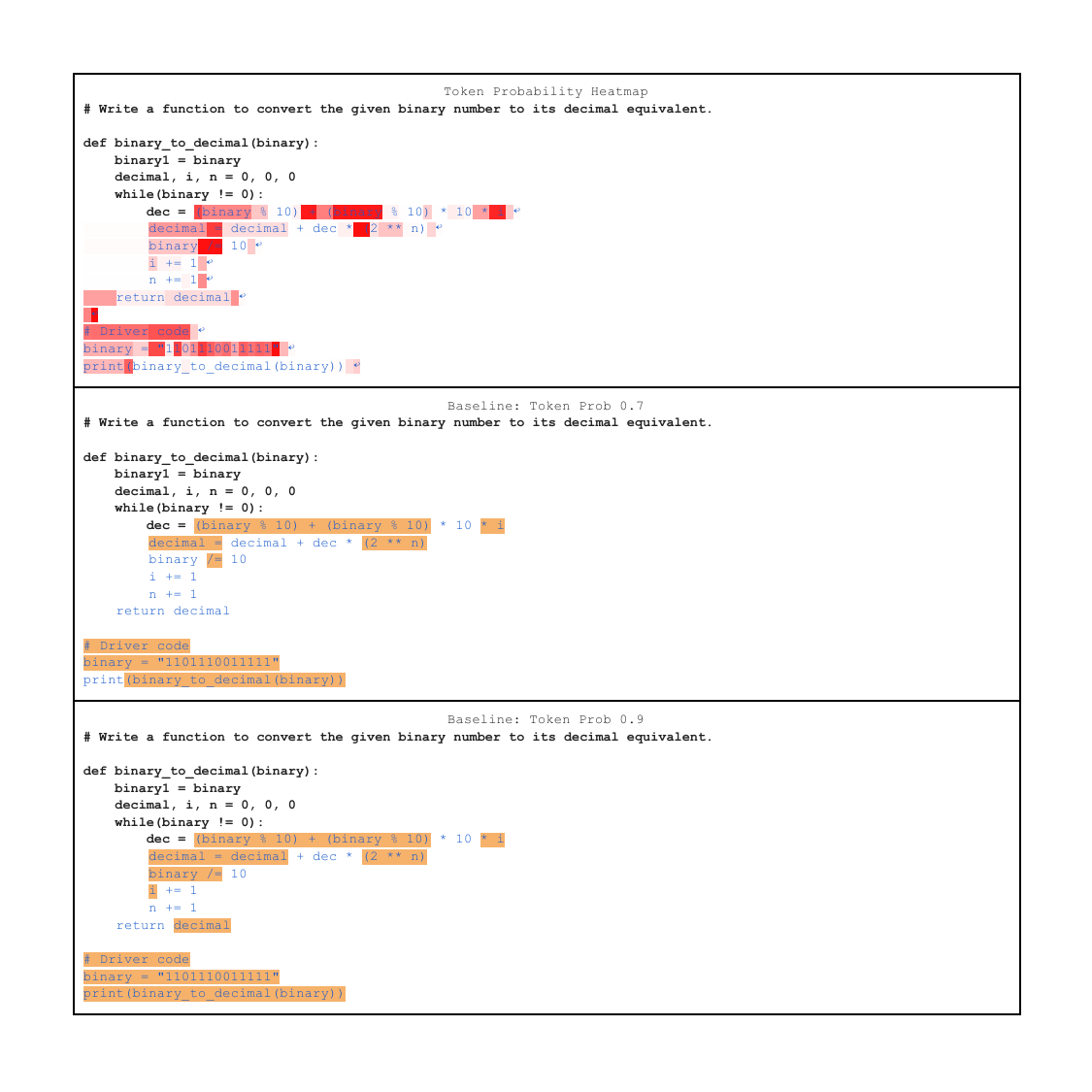}
    \caption{Per-token conditional probability heatmap and output of token-probability-based baselines for the MBPP example in \cref{fig:app:example3}. Note that low-conditional-prob. tokens (such as the `\texttt{/=}' after `\texttt{binary}`) are frequently followed by high-conditional-prob. tokens that only make sense in context of the earlier tokens (such as `\texttt{10}').}
    \label{fig:app:example4}
\end{figure}

\begin{figure}[p]
    \centering
    \includegraphics[page=8,trim=0.5in 5in 0.5in 0.5in,clip,width=\linewidth]{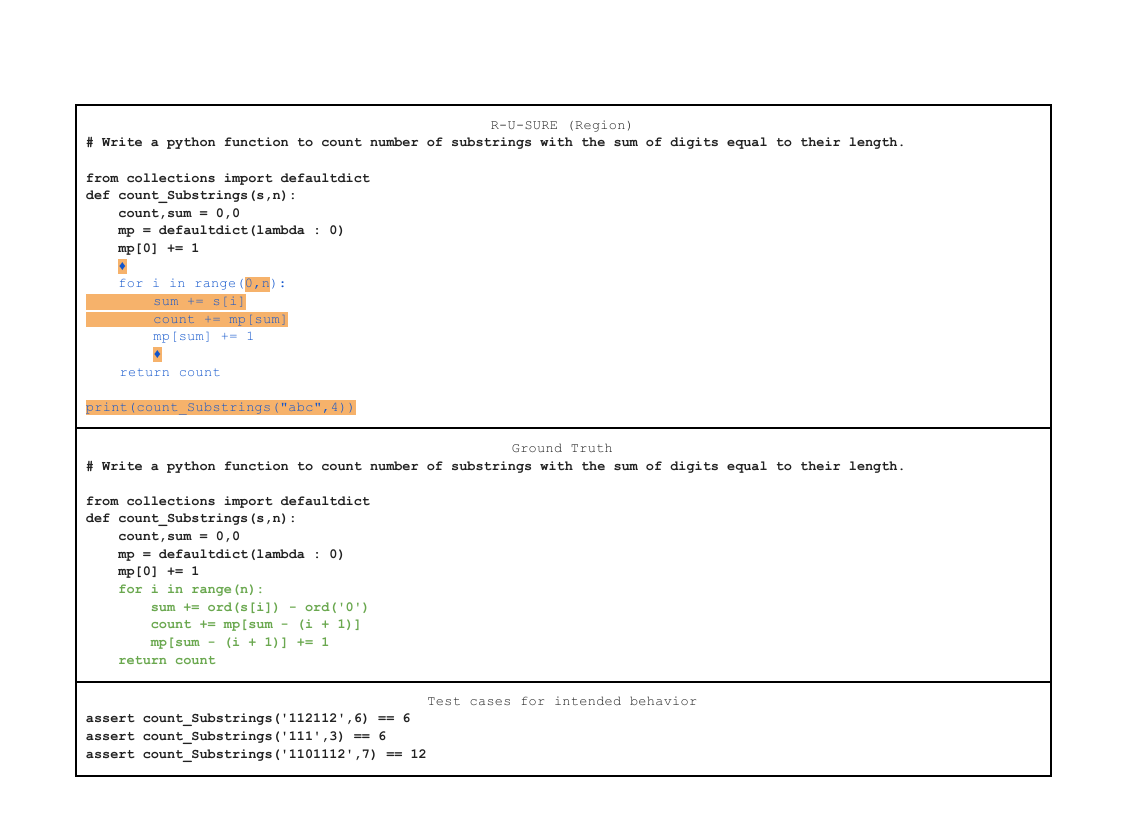}
    \caption{Output of R-U-SURE compared to the ground truth for another example in the Mostly Basic Python Problems benchmark dataset \citep{Austin2021ProgramSW}. We manually selected a location in the MBPP reference solution, then fed the prefix to the model. Again, the model's implementation does not exactly match the intended behavior. In this case, most incorrect parts are highlighted, but there are some changes that must also be made outside of highlighted regions.
    The `$\blacklozenge$' character denotes a location where additional statements might be inserted.
    }
    \label{fig:app:example5}
\end{figure}

\begin{figure}[p]
    \centering
    \includegraphics[page=9,trim=0.5in 4in 0.5in 0.5in,clip,width=\linewidth]{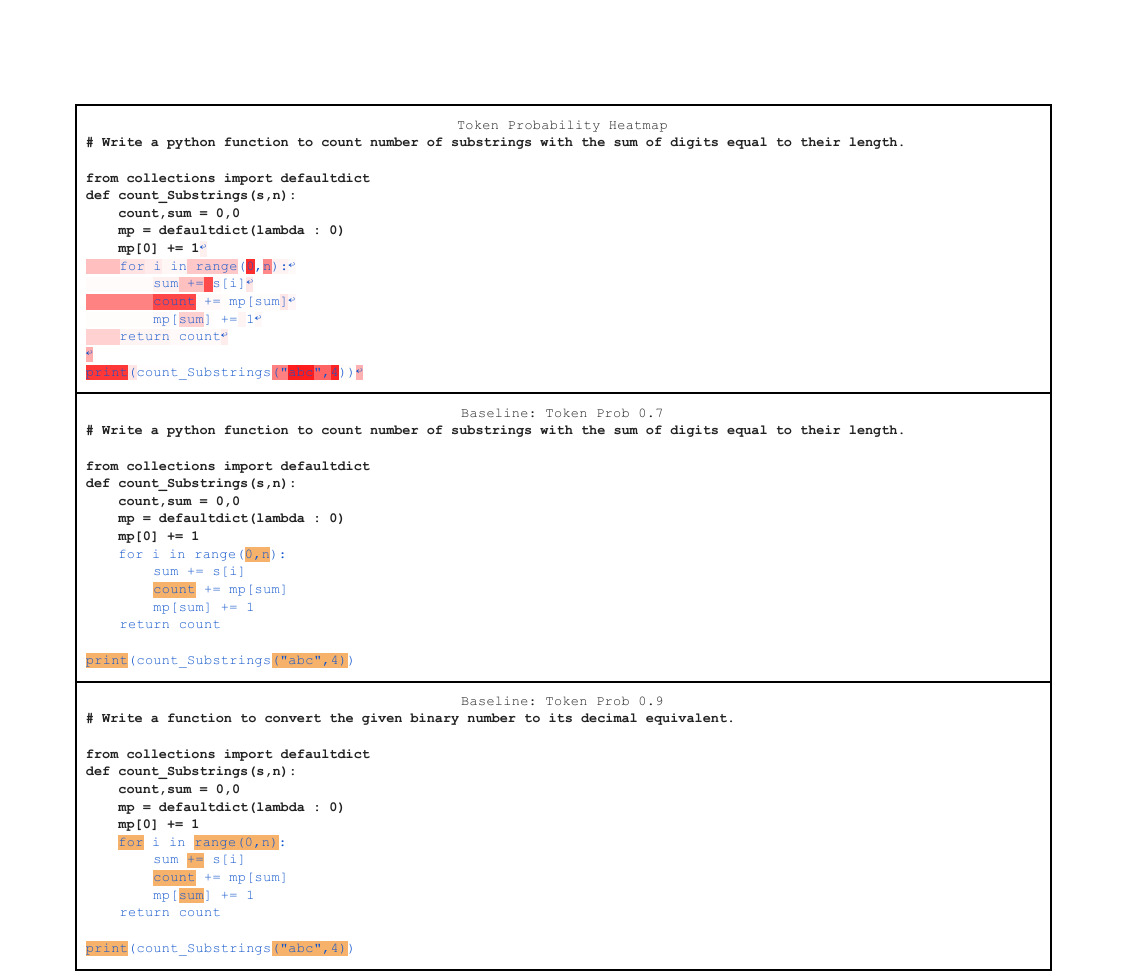}
    \caption{Per-token conditional probability heatmap and output of token-probability-based baselines for the MBPP example in \cref{fig:app:example5}.}
    \label{fig:app:example6}
\end{figure}

\begin{figure}[p]
    \centering
    \includegraphics[page=10,trim=0.5in 4.5in 0.5in 0.5in,clip,width=\linewidth]{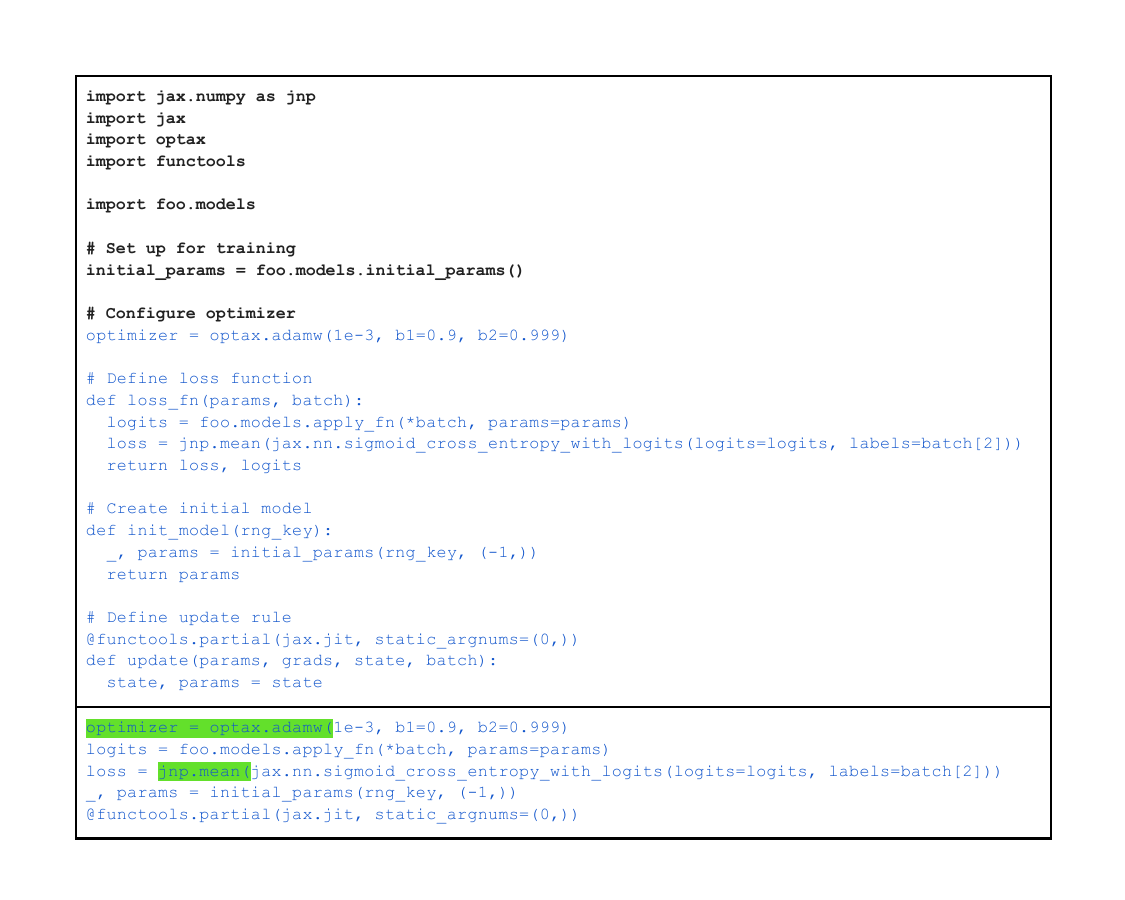}
    \caption{Output of R-U-SURE (Useful Calls) for a handwritten prompt involving usage of \texttt{optax}. After postprocessing, the only calls that appear often enough in the model samples to be extracted are calls to \texttt{optax.adamw} and \texttt{jnp.mean}; these would be good candidates for preemptively showing documentation.}
    \label{fig:app:example7}
\end{figure}

\subsection{Runtime of R-U-SURE}\label{app:examples-runtime}

The wall-clock runtime of our implementation of the R-U-SURE system depends on the number of samples as well as the complexity of the programs. We demonstrate this by measuring the runtime for the two prompts shown in \cref{fig:app:example1} and \cref{fig:app:example3}, across varying number of model samples and sample lengths. On a GCP n1-standard-8 virtual machine, we obtain the following results:

\begin{itemize}
\item Combining eight model samples, each restricted to eight lines, takes about 20 to 60 milliseconds for the dual decomposition solver and about 1 to 1.5 seconds for the parsing and diagram construction logic.
\item Combining 32 eight-line samples takes between 0.1 and 1.5 seconds for the solver and about 3 to 5 seconds for parsing/diagram construction.
\item Combining 32 longer model samples (with 256 vocabulary tokens, or about 23 lines) can take between 0.5 and 6 seconds for the solver and between 8 and 40 seconds for parsing/diagram construction depending on complexity (with the example in \cref{fig:app:example1} taking the longest).
\end{itemize}

We note that, in our current implementation, the parsing/diagram construction logic is designed to be flexible and makes heavy use of Python dictionaries. This could likely be sped up considerably for a specialized application. The solver can also be interrupted if necessary to obtain a possibly-suboptimal solution in a fixed amount of time.

In terms of asymptotic complexity, the time and space required to build the system and each iteration of coordinate ascent scales as $O(\ell^2 K)$, where $\ell$ is the length of the model suggestions and $K$ is the number of samples. 
\clearpage
\section{Decision Diagrams: Definitions and Algorithms}\label{app:decision-diagram-definitions}

In this section, we discuss our definition of decision diagrams and describe how we use them to enable efficient algorithms.

\subsection{Our Definitions}

\begin{definition}\label{defn:bdd} A \textbf{(nondeterministic, weighted) binary decision diagram} (BDD) $D$ over binary vectors $\vb \in \{0,1\}^d$ is a directed acyclic graph consisting of
\begin{itemize}
    \item a node set $N$,
    \item an arc set $A$,
    \item mappings $h : A \to N$ and $t : A \to N$ such that each arc $a$ is directed from node $h(a)$ to node  $t(a)$,
    \item a mapping $w :  A \to \R$ such that $w(a)$ is the weight of arc $a$,
    \item a mapping $\alpha : A \to (\{1, \dots, d\} \times \{0,1\}) \sqcup \{\textsc{None}\}$ such that, if $\alpha(a) = (i, \beta)$, then this edge can only be used when $\vb_i = \beta$, and if $\alpha(a) = \textsc{None}$, this edge can always be used. 
    \item a source node $\top \in N$, which is not the tail of any arc,
    \item a sink node $\bot \in N$, which is not the head of any arc.
\end{itemize}
\end{definition}

\begin{definition}\label{defn:computation-path} A \textbf{computation path} for a binary vector $\vb \in \{0,1\}^d$ is a sequence of arcs $P = (a_1, a_2, \dots, a_n)$ from $\top$ to $\bot$ that are consistent with $\vb$, e.g. such that
$h(a_1) = \top$, $h(a_{i+1}) = t(a_i)$ for $1 \le i < n$, $t(a_n) = \bot$, and if $\alpha(a_i) = (j, \beta)$ for any $i$ then $\vb_j = \beta$. The weight of this path is the sum of arc weights $\sum_i w(a_i)$, which by abuse of notation we will denote $w(P)$. We denote the set of all computation paths for a particular vector $\vb$ as $\cP(D, \vb)$.
\end{definition}

\begin{definition}\label{defn:represents} A BDD $D$ \textbf{represents} a binary function $w : \{0,1\}^d \to \R \cup \{-\infty\}$ (under $\max$-aggregation) if, for all $\vb \in \{0,1\}^d$, we have
\[
w(\vb) = \max_{P \in \cP(D, \vb)}w(P),
\]
e.g. this is the weight of the maximum-weight path from $\top$ to $\bot$ consistent with $\cP$, or $-\infty$ if there are no such paths.
\end{definition}

\begin{definition}\label{defn:ordered-bdd} A BDD $D$ is \textbf{ordered} if its nodes can be partitioned into layers according to some partition function $\ell : N \to \{0, 1, \dots, d\}$ such that $\ell(\top) = 0$, $\ell(\bot) = d$, and for each arc $a \in A$:
\begin{itemize}
    \item if $\alpha(a) = \textsc{None}$, then $\ell(h(a)) = \ell(t(a))$,
    \item if $\alpha(a) = (i, \beta)$, then $\ell(h(a)) = i-1$ and $ \ell(t(a)) = i$.
\end{itemize}
Intuitively, an ordered BDD is a BDD such that any path from $\top$ to $\bot$ assigns every index of $\vb$ exactly once, in order of increasing index.
\end{definition}
\begin{definition} A \textbf{system of BDDs} is a collection of BDDs $D_i$ over the same set of binary vectors $\vb \in \{0,1\}^d$. We say that a system of BDDs represents a binary function $w : \{0,1\}^d \to \R \cup \{-\infty\}$ if $w$ can be written as a sum
\[
w(\vb) = \sum_i w^{(i)}(\vb)
\]
and $D_i$ represents $w^{(i)}$ for each $i$.
\end{definition}

Our approach described in \cref{sec:representing-as-decision-diagrams} can now be described more specifically as rewriting our original objective using a set of binary functions
\[
w^{(k)}(\vb) = \begin{cases}
\frac{1}{K} u(g^{(k)}, f(\vb)) &\vb \in \cB,\\
-\infty &\text{otherwise}.
\end{cases}
\]
and then representing each such function with an ordered BDD. More generally, we allow representing $w^{(k)}(\vb)$ as a system of BDDs $(D^{(k)}_1, D^{(k)}_2, \dots, D^{(k)}_{m})$, and take advantage of this flexibility to efficiently separate the computation of the utility function from constraints, which we describe in more detail in \cref{app:utility-functions}. Combining all of the BDDs or BDD systems for each value of $k$ then yields a system $(D_1, \dots, D_J)$ that represents the total utility
\[
U(\vb) = \sum_{k=1}^K w^{(k)}(\vb)
\]
as estimated across the samples $g^{(1)}, \dots, g^{(k)}$. (Note that the number $K$ of samples may or may not match the number of decision diagrams $J$ in general, depending on whether any of the $w^{(k)}$ were represented as more than one diagram).

We note that any function $w^{(k)} : \cB \to \R$ can be expressed as a weighted binary decision diagram, but the size of the diagram may grow exponentially with the number of binary choices $d$ \citep{hooker2013decision}. However, our edit-distance-based utility functions $u(g,s)$ can be represented as decision diagram whose size grows only quadratically with the number of tokens in $s$ and $g$, due to the similarity between decision diagrams and dynamic programming algorithms.

\subsection{A comparison to other definitions of decision diagrams}
Decision diagrams have seen a number of uses for a variety of combinatorial optimization and search problems; \citet{Castro2022DecisionDF} gives an overview of many such uses.
Here we briefly summarize some of the differences between our definition and others in the literature.

\paragraph{Determinism} Many definitions of decision diagrams (e.g. \citet{lozano2020consistent,lange2021efficient}) focus on \emph{deterministic} decision diagrams, which have the additional properties that
\begin{itemize}
    \item every node $n$ other than $\bot$ is associated with a particular decision variable $\vb_{\text{var}(n)}$ with $\text{var}(n) \in \{1, \dots, d\}$,
    \item there are at most two edges from any given node $n$ (i.e. with $h(a) = b$): one which assigns $\alpha(a) = (\text{var}(n),  0)$ and one which assigns $\alpha(a) = (\text{var}(n),  0)$,
    \item every arc assigns some variable, e.g. there is no edge with $\alpha(a) = \textsc{None}$.
\end{itemize}
Nondeterministic decision diagrams are related to deterministic ones in the same way that nondeterministic finite automata relate to deterministic finite automata: for a deterministic decision diagram, you can read off a single computation path $P$ for a given vector $\vb$ if it exists by following the sequence of branches, whereas for a nondeterministic decision diagram, you may need to search over many consistent sub-paths to identify one or more computation paths for a specific vector.

Some definitions of nondeterministic decision diagrams define them by introducing two types of node: ordinary nodes, which are associated with variables have two outgoing arcs tagged 0 and 1, and nondeterministic nodes, which have no variable and any number of outgoing arcs with $\alpha(a) = \textsc{None}$ \citep{Bollig2018OnTR}. For simplicity, our definition does not directly constrain edges based on any assignment of nodes to decision variables, but the two formulations are equivalently expressive, especially for ordered nondeterministic BDDs (for which $\ell$ approximately corresponds to a node-variable association).

\paragraph{Ambiguity} Most definitions of nondeterministic BDDs focus on \emph{unambiguous} nondeterministic BDDs, for which there is at most one computation path for any binary vector $\vb$ \citep{Bollig2018OnTR}; these can also be referred to as \emph{exactly representing} specific binary functions \citep{Castro2022DecisionDF}. In contrast, we explicitly allow BDDs to be ambiguous, and resolve conflicts by taking the max over edges. This makes it significantly easier to express our utility functions as decision diagrams, by essentially interleaving the edit-distance search algorithm with the decision diagram as part of a single optimization problem.

It turns out to be very straightforward to extend the min-(or max-)marginal averaging technique of \citet{lange2021efficient} to work for ambiguous decision diagrams with only minimal changes, as we describe in the next section.

\paragraph{Reduction} A common method for obtaining more efficient representations of decision diagrams is to reduce them to a particular canonical form, collapsing nodes that serve identical roles \citep{hooker2013decision,Castro2022DecisionDF}. While it may be possible to reduce our decision diagrams to a more efficient form, we do not attempt to produce reduced decision diagrams in our implementation.

\paragraph{Binary v.s. multivalued} Some definitions of decision diagrams allow variables to be assigned to values in a larger finite set $\cV$; these are known as multivalued decision diagrams \citep{hooker2013decision,Castro2022DecisionDF}. In practice, we implement our utility functions as multivalued decision diagrams; however, to make derivations simpler for the Lagrangian relaxation, we encode these multivalued choices as one-hot-encoded binary vectors before running our max-marginal optimization process.

\subsection{Efficient algorithms for max-marginal message passing on BDDs}
We now describe how to efficiently optimize a Lagrangian relaxation of a BDD system, as described in \cref{subsec:dual-decomposition-decision-diagrams,sec:representing-as-decision-diagrams}. We consider the objective
\begin{align}
U = \max_{\vb \in \{0, 1\}^{d} } \sum_{j=1}^J w^{(j)}(\vb),
\end{align}
where we have changed the indexing to account for situations where the number of decision diagrams $J$ does not equal the number of model samples $K$. We then construct the Lagrangian relaxation
\begin{align}
W^{(j)}(\vb^{(j)}, \vlambda) = w^{(j)}(\vb^{(j)}) + \vlambda^{(j)} \cdot \vb^{(j)},
\\
L(\vlambda^{(1:J)}) = \sum_{j=1}^J \max_{\vb^{(j)}} W^{(j)}(\vb^{(j)}, \vlambda)
\end{align}
where we require that $\sum_m \vlambda^(j) = 0$. Intuitively, if we have $\vb^{(j)}_i \ne \vb^{(j')}_i$, we can adjust the $\vlambda^{(j)}_i$ and $\vlambda^{(j')}_i$ in opposite directions to remove any utility benefits of violating the equality constraint. (However, this penalty acts independently on each variable, and may not be able to simultaneously enforce agreement for joint configurations of variables; this is what causes a nonzero duality gap.) We note that this is the same relaxation described by \citet[Section 3.1]{lange2021efficient}, except for the more general form of $w^{(j)}$ (not just for linear programs) and the use of $\max$ rather than $\min$.

The max-marginal coordinate ascent update with respect to the variable block $(\lambda^{(1)}_i, \lambda^{(2)}_i, \dots, \lambda^{(K)}_i)$, as derived by \citet{lange2021efficient}, is then given by
\begin{align}
m^{(j)}_{i := \beta} &= \max_{\substack{\vb^{(j)} ~\text{s.t.}~ \vb^{(j)}_i = \beta}} W^{(j)}(\vb^{(j)},\vlambda), \qquad \beta \in \{0, 1\}\\
\vlambda^{(j)}_i &\gets \vlambda^{(j)}_i - (m^{(j)}_{i := 1} - m^{(j)}_{i := 0}) + \frac{1}{J} \sum_{j'}\left(m^{(j')}_{i := 1} - m^{(j')}_{i := 0}\right).
\label{eqn:app-min-marginal-update-binary}
\end{align}
Our main requirement for computing this update is that we can efficiently compute the $m^{(j)}_{i := \beta}$ for our current values of $\vlambda$. Fortunately, this can be done for nondeterministic weighted BDDs using a straightforward dynamic programming algorithm. This algorithm maintains two cached dynamic programming tables (\textsc{Prefix} and \textsc{Suffix}) in order to make updates efficient: \textsc{Prefix} stores the maximum weight from $\top$ to a given node (sorted by level $\ell$), and  \textsc{Suffix} stores the maximum weight from each node to $\bot$. We initialize these tables using \cref{alg:bdd-table-initialization}, then run \cref{alg:bdd-max-marginal} to compute desired max marginals. Then, each time we update values for $\vlambda_i^{(1:J)}$, we must invalidate the caches for index $i$ by running \cref{alg:bdd-cache-invalidation}.

A key property of this algorithm is that modifying the dual variables for a particular decision variable $\vb^{(j)}_i$ only affects prefixes and suffixes that include assignments to $\vb^{(j)}_i$. Thus, if we wish to compute max-marginals for $\vb^{(j)}_{i-1}$ or $\vb^{(j)}_{i+1}$ next, we can reuse almost all of the values from the cache, and only update the prefixes that changed due to modifications to $\vlambda_i^{(j)}$.

We take advantage of this property by running a series of alternating forward and backward sweeps,  updating $\vlambda_1^{(1:J)}, \vlambda_2^{(1:J)}, \dots, \vlambda_d^{(1:J)}$ during a forward sweep and then $\vlambda_d^{(1:J)}, \vlambda_{d-1}^{(1:J)}, \dots, \vlambda_1^{(1:J)}$ in a backward sweep. Each of these sweeps visits every arc twice (once to compute max marginals and once to update the modified prefix or suffix), enabling us to run an entire min-marginal-averaging cycle with time complexity proportional to the size of the decision diagram.

Note that this algorithm is not guaranteed to find a primal solution if there is a nonzero dual gap, and may get stuck in certain fixed points even if the dual gap is zero \citep{werner2020relative}. In our experiments, however, we find that the bound is tight (to within machine precision) over 85\% of the time.

\begin{algorithm}[t]
\caption{Stateful dynamic programming algorithm initialization step}
\label{alg:bdd-table-initialization}
\begin{algorithmic}
\STATE {\bfseries Input:} BDD $D_j = (N, A, h, t, w, \alpha, \top, \bot)$ with order $\ell$, caches \textsc{Prefix} and \textsc{Suffix}
\STATE\COMMENT{\# Compute initial prefixes}
\STATE Initialize $\textsc{Prefix}[0, \top] = 0.0$
\FOR{each arc $a \in A$ with $\ell(h(a)) = \ell(t(a)) = 0$), in topologically-sorted order}
    \STATE Set $\textsc{Prefix}[k, t(a)] = \max(\textsc{Prefix}[k, t(a)], \textsc{Prefix}[k, h(a)] + w(a))$
\ENDFOR
\STATE\COMMENT{\# Compute initial suffixes}
\STATE Initialize $\textsc{Suffix}[d, \top] = 0.0$
\FOR{each arc $a \in A$ with $\ell(h(a)) = \ell(t(a)) = d$), in reverse topologically-sorted order}
    \STATE Set $\textsc{Suffix}[k, h(a)] = \max(\textsc{Suffix}[k, h(a)], \textsc{Suffix}[k, t(a)] + w(a))$
\ENDFOR
\end{algorithmic}
\end{algorithm}

\begin{algorithm}[t]
\caption{Stateful dynamic programming cache invalidation step}
\label{alg:bdd-cache-invalidation}
\begin{algorithmic}
\STATE {\bfseries Input:} BDD $D_j = (N, A, h, t, w, \alpha, \top, \bot)$ with order $\ell$, updated index $i$, caches \textsc{Prefix} and \textsc{Suffix}
\FOR{$k$ {\bfseries in} $[i, i+1, \dots, d]$}
    \STATE Delete all entries of $\textsc{Prefix}[k, :]$
\ENDFOR
\FOR{$k$ {\bfseries in} $[0, 1, \dots, i-1]$}
    \STATE Delete all entries of $\textsc{Suffix}[k, :]$
\ENDFOR
\end{algorithmic}
\end{algorithm}

\begin{algorithm}[p]
\caption{Stateful dynamic programming algorithm for $m^{(j)}_{i := \beta}$}
\label{alg:bdd-max-marginal}
\begin{algorithmic}
\STATE {\bfseries Input:} BDD $D_j = (N, A, h, t, w, \alpha, \top, \bot)$ with order $\ell$, desired variable index $i$, $\vlambda^{(j)}$, caches \textsc{Prefix} and \textsc{Suffix}
\STATE\COMMENT{\# Compute necessary prefixes}
\FOR{$k$ {\bfseries in} $[1, 2, \dots, i-1]$}
    \IF{\textsc{Prefix} does not have values for level $k$}
        \FOR{each node $n \in N$ with $\ell(n) = k$}
            \STATE Initialize $\textsc{Prefix}[k, n] = -\infty$
        \ENDFOR
        \STATE\COMMENT{\# Process edges that assign $\vb_k$}
        \FOR{each arc $a \in A$ with $\ell(h(a)) = k-1$ and $\ell(t(a)) = k$, in topologically-sorted order}
            \STATE Let $(v, \beta) = \alpha(a)$, assert $v = k$ \COMMENT{~~\# $a$ must assign to $\vb_k$ by \cref{defn:ordered-bdd}}
            \IF[~~\# Need to perturb by $\vlambda^{(j)}_k$]{$\beta = 1$}
            \STATE Set $\textsc{Prefix}[k, t(a)] = \max(\textsc{Prefix}[k, t(a)], \textsc{Prefix}[k-1, h(a)] + w(a) + \vlambda^{(j)}_k)$
            \ELSE
            \STATE Set $\textsc{Prefix}[k, t(a)] = \max(\textsc{Prefix}[k, t(a)], \textsc{Prefix}[k-1, h(a)] + w(a))$
            \ENDIF
        \ENDFOR
        \STATE\COMMENT{\# Process edges in level $k$}
        \FOR{each arc $a \in A$ with $\ell(h(a)) = k$ and $\ell(t(a)) = k$, in topologically-sorted order}
            \STATE Set $\textsc{Prefix}[k, t(a)] = \max(\textsc{Prefix}[k, t(a)], \textsc{Prefix}[k, h(a)] + w(a))$
        \ENDFOR
    \ENDIF
\ENDFOR
\STATE\COMMENT{\# Compute necessary suffixes}
\FOR{$k$ {\bfseries in} $[d-1, d-2, \dots, i]$}
    \IF{\textsc{Suffix} does not have values for level $k$}
        \FOR{each node $n \in N$ with $\ell(n) = k$}
            \STATE Initialize $\textsc{Suffix}[k, n] = -\infty$
        \ENDFOR
        \STATE\COMMENT{\# Process edges that assign $\vb_{k+1}$}
        \FOR{each arc $a \in A$ with $\ell(h(a)) = k$ and $\ell(t(a)) = k + 1$, in reverse topologically-sorted order}
            \STATE Let $(v, \beta) = \alpha(a)$, assert $v = k + 1$ \COMMENT{~~\# $a$ must assign to $\vb_{k+1}$ by \cref{defn:ordered-bdd}}
            \IF[~~\# Need to perturb by $\vlambda^{(j)}_{k + 1}$]{$\beta = 1$}
            \STATE Set $\textsc{Suffix}[k, h(a)] = \max(\textsc{Suffix}[k, h(a)], \textsc{Suffix}[k + 1, t(a)] + w(a) + \vlambda^{(j)}_{k + 1})$
            \ELSE
            \STATE Set $\textsc{Suffix}[k, h(a)] = \max(\textsc{Suffix}[k, h(a)], \textsc{Suffix}[k + 1, t(a)] + w(a))$
            \ENDIF
        \ENDFOR
        \STATE\COMMENT{\# Process edges in level $k$}
        \FOR{each arc $a \in A$ with $\ell(h(a)) = k$ and $\ell(t(a)) = k$, in reverse topologically-sorted order}
            \STATE Set $\textsc{Suffix}[k, t(a)] = \max(\textsc{Suffix}[k, t(a)], \textsc{Suffix}[k, h(a)] + w(a))$
        \ENDFOR
    \ENDIF
\ENDFOR
\STATE\COMMENT{\# Compute max marginals}
\STATE Initialize $m^{(j)}_{i := 0}$ and $m^{(j)}_{i := 1}$ to $-\infty$
\FOR{each arc $a \in A$ with $\ell(h(a)) = i - 1$ and $\ell(t(a)) = i$}
    \STATE Let $(v, \beta) = \alpha(a)$, assert $v = i$ \COMMENT{~~\# $a$ must assign to $\vb_{i}$ by \cref{defn:ordered-bdd}}
    \STATE Set $m^{(j)}_{i := \beta} = \max(m^{(j)}_{i := \beta},  \textsc{Prefix}[i - 1, h(a)]+ w(a) + \textsc{Suffix}[i, t(a)] + \vlambda^{(j)}_{i})$
\ENDFOR
\RETURN $m^{(j)}_{i := 0}, m^{(j)}_{i := 1}$
\end{algorithmic}
\end{algorithm}

\subsection{Extension to multivalued decision diagrams}
In practice, although we analyze and implement our algorithms as if we are optimizing over binary variables, it is more convenient for our utility functions to be written in terms of assignments to an arbitrary finite set of values $\cV$; this is sometimes known as a ``multivalued'' decision diagram \citep{hooker2013decision}. We do this by enumerating the values of $\cV$, and treating a particular choice $x_i = v \in \cV$ as a collection of ``indicator'' assignments $\vb_{(i, v)} := 1$, $\vb_{(i, v')} := 0$ for $v' \ne v$.

We take advantage of our knowledge of this indicator structure when running our max-marginal step, to simplify the implementation. In particular, we perform simultaneous block updates over all indicator variables, computing
\begin{align}
m^{(j)}_{(i, v) := 1} &= \max_{\substack{\vb^{(j)} ~\text{s.t.}~ \vb^{(j)}_{(i,v)} = 1,}} W^{(j)}(\vb^{(j)},\vlambda)
\\
\vlambda^{(j)}_{(i, v)}
&\gets
\vlambda^{(j)}_{(i, v)}
- m^{(j)}_{{(i, v)} := 1}
+ \frac{1}{J} \sum_{j'} m^{(j')}_{{(i, v)} := 1}.\label{eqn:app-min-marginal-update-multivalued}
\end{align}
which is the update from \cref{eqn:app-min-marginal-update-binary} but dropping the $m^{(j)}_{{(i, v)} := 0}$ terms. This works because we know that, for any valid assignment to the indicator variables, exactly one such indicator will be active. Thus, each of the $m^{(j)}_{{(i, v)} := 0}$ terms is equal to $m^{(j)}_{{(i, v')} := 1}$ for some alternative assignment $v'$, which means making the $m^{(j)}_{{(i, v')} := 1}$ agree is sufficient to make the differences $m^{(j)}_{{(i, v)} := 1} - m^{(j)}_{{(i, v)} := 0}$ agree as well. (Indeed, in our actual implementation of \cref{alg:bdd-max-marginal}, we do not bother computing entries for the $m^{(j)}_{{(i, v)} := 0}$ at all, since they are unused in the update \cref{eqn:app-min-marginal-update-multivalued}.)

This indicator representation also allows us to reuse parts of our implementation when decoding a heuristic primal solution, in the situations where our solver fails to find a setting for the dual variables that makes the dual bound tight. Specifically, we iterate through all of the variables, and greedily select the best assignment
\[
v_i^* = \argmax m^{(j)}_{(i, v) := 1}
\]
then set
\[
\vlambda^{(j)}_{(i, v')} \gets -\infty
\]
for each $v' \ne v_i^*$. This effectively prunes any arc that assigns a different value from the graph, ensuring we decode a single consistent assignment.

\clearpage
\newpage
\section{Utility functions and tree representation}\label{app:utility-functions}

In this section, we give a high level description of our utility function implementation and of the tree representation we use for combining suggestions. We will also include the code for our utility functions in a later open-source release.

\subsection{Tree representation}

We represent the model samples and user intents as possibly-nested sequences of nodes of the following types:

\begin{itemize}
    \item \textbf{Token nodes} represent programming language tokens, which we should try to match between the suggestion and the target intent. Token nodes contain a source string and optionally a type, and any two nodes with the same string and the same type will match. We typically use the type to encode information about the AST nodes.
    \item \textbf{Decoration nodes} denote locations of whitespace or other aspects of the suggestion that do not need to be considered as part of the edit distance calculation. These are not used during optimization.
    \item \textbf{Group nodes} contain an arbitrary number of child nodes, which may be token nodes, decoration nodes, or other group nodes. Each group node has an optional type, and any two group nodes of the same type can be matched together; matching two group nodes involves running an edit distance calculation on their children subsequences.
\end{itemize}

The suggestion prototype, usually the model sample with the highest probability, is augmented with a few additional nodes:

\begin{itemize}
    \item \textbf{Region start nodes} represent locations where we may start a confidence region.  Depending on the configuration, such regions may represent pockets of \LowConf{} within a default of \HighConf{} (e.g. for detecting edit locations), or pockets of \HighConf{} within a default of \LowConf{} (e.g. for extracting a subsequence of API calls).
    \item \textbf{Region end nodes} represent locations where we may end a confidence region that we started earlier in the (sub)sequence. Note that every confidence region that starts inside a group node is required to end within that same group node.
    \item \textbf{Truncation nodes} represent locations where we may decide to truncate the suggestion.
\end{itemize}

These nodes are inserted in various locations into the parse tree with a preprocessing step, which gives us a large amount of control over the space of augmented suggestions $\cS$. For instance, for the edit localization task, we do not allow \LowConf{} regions to include single parentheses or brackets by placing matched brackets into a group and not allowing regions to start or end at the boundary of those groups. For the API call task, we use the region start/end nodes to identify \HighConf{} calls, but only allow calls to be selected one at a time by only inserting them inside the relevant call groups.

\subsection{Utility function}
We now describe our base utility function at a high level; the specific applications are determined by configuring this utility function with different costs and constraints.

\subsubsection{Utility Configuration}
Our utility function implementation is configured by a set of edit penalties:
\begin{itemize}
    \item For each confidence level:
    \begin{itemize}
        \item A per-character or per-token utility for matching tokens in the suggestion with those in the ground truth,
        \item A per-character or per-token cost for deleting tokens in the suggestion
        \item A penalty for starting to edit (either inserting or deleting)
    \end{itemize}
    \item A penalty for changing confidence levels (e.g. to encourage fewer blocks of \LowConf{}).
\end{itemize}

\subsubsection{Edit-Based Decision Diagram}
Nodes in our decision diagram (which we call ``states'' to distinguish them from tree nodes, by analogy to finite state machines) are associated with a tuple of positions, one in the prototype and one in the hypothetical target intent (usually generated from the model), in a similar way as in \cref{alg:simple-edit-distance-decision-diagram}.

We further group our states into a number of types, used to track the progress of edits.
The list of state types are:
\begin{itemize}
    \item \textbf{PROCESS-PROTOTYPE (ADVANCE)}: We are advancing past region start/end nodes or truncation nodes in the prototype. We can either stay in PROCESS-PROTOTYPE and move past one of those nodes in the prototype, or transition to MATCH to match tokens or groups, or transition to MAY-DELETE if we need to edit at this location, which incurs an additional penalty.
    \item \textbf{MAY-DELETE}: We have decided we need to make an edit at this location. We are allowed to delete an arbitrary number of the prototype; we may also process any region start/end nodes or truncation nodes we see. We then transition into MAY-INSERT.
    \item \textbf{MAY-INSERT}: We are allowed to insert an arbitrary number of nodes in the target. We always insert after deleting, to reduce the number of redundant paths in the graph. Once we have inserted all that we need to, we can transition to MATCH.
    \item \textbf{MATCH}: We are prepared to match nodes in the prototype and target, after which we return to PROCESS-PROTOTYPE; we can also end the subproblem if we are at the end of two group nodes.
    \item \textbf{RECURSIVELY-DELETING (FORCED)}: We have committed to deleting an entire subtree, and are now deleting each of the nodes in it. We cannot stop until we exit the subtree.
    \item \textbf{RECURSIVELY-INSERTING (FORCED)}: We have committed to inserting an entire subtree, and are now inserting each of the nodes in it. We cannot stop until we exit the subtree.
\end{itemize}

 Additionally, each node is associated with a confidence level (\HighConf{} or \LowConf{}); the active confidence level determines the utility associated with each of the state transitions described above.
 
Token nodes are handled depending on the state; in MATCH we must align two identical tokens to proceed, whereas in MAY-DELETE or MAY-INSERT we are allowed to delete or insert tokens individually.

Group nodes are handled using a recursive call. If we are processing two group nodes and we are in the MATCH state, we recursively build a decision diagram for the subsequences of the two nodes. If we delete a group node in the MAY-DELETE state, we call a recursive helper function that builds a small decision diagram that only deletes nodes and stays in the RECURSIVELY-DELETING state. Inserts are handled in an analogous way.

We implicitly embed the space of suggestions $\cS(g^{(1)})$ into a space of binary vectors by introducing decisions for each of the control nodes. Here we focus on the version of our task that introduces \LowConf{} regions into a suggestion.
\begin{itemize}
    \item At a Region Start node, if we are currently in \HighConf{}, we can transition to \LowConf{}. We track this choice with a decision variable assignment.
    \item At a Region End node, if we are currently in \LowConf{}, we can transition to \HighConf{}. We track this choice with a decision variable assignment.
    \item At a truncation node, we can choose to immediately jump from our current state to the final state, paying no more penalties but receiving no additional reward. We track this choice with a decision variable assignment.
\end{itemize}
We additionally include decision variables that track whether each token was inside a annotated region when we processed it; this information is redundant with the start/end nodes, but can improve the optimization by providing additional information in the message passing iterations. We then order these decision variables by their order of appearance in the graph, and interpret the values of each decision as the embedding $\phi(s)$ of each possible suggestion.

\begin{sidewaysfigure}[p]
    \centering
    \includegraphics[width=\textwidth,trim=0.3in 6.5in 0.3in 0,clip]{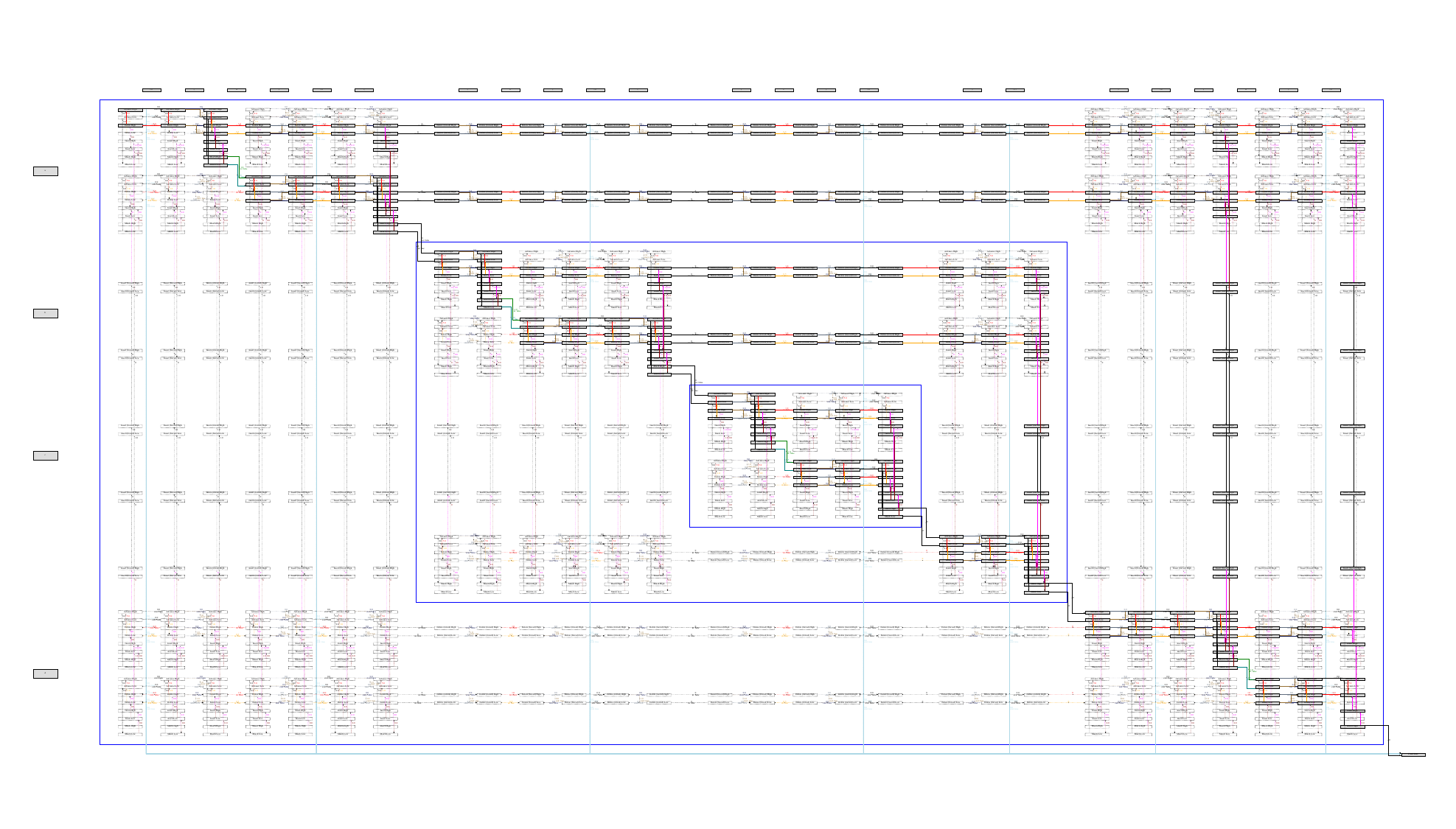}
    \caption{Rendering of the edit decision diagram with all features enabled, for aligning the sequence ``a [ b [ c ] ] d'' with itself.}
    \label{fig:app:edit-dag}
\end{sidewaysfigure}

\cref{fig:app:edit-dag} shows a rendering of the decision diagram we construct when combining two simple sequences. Note that the diagrams we use to combine actual model samples are much larger, since every token of the suggestion is represented by multiple states in the diagram. Also, this diagram is written in terms of negative utility (e.g. as a collection of costs).

\begin{figure}[t]
    \centering
    \includegraphics[width=\textwidth,trim=3.6in 8.25in 3.2in 1.75in,clip]{figures/appendix/edit-dag.pdf}
    \caption{Zoomed-in view of a portion of the edit decision diagram in \cref{fig:app:edit-dag}}
    \label{fig:app:edit-dag-zoom}
\end{figure}

\subsubsection{Constraint Decision Diagram}
The above decision diagram ensures that edits respect the tree structure, but does not by itself ensure that annotated regions are aligned with that tree structure. We address this by building a second decision diagram, which depends only on the prototype sequence and which enforces the constraints on the annotated regions.

The second DAG tracks a more fine-grained set of confidence types:
\begin{itemize}
    \item \textbf{OUTSIDE-REGION}: We are outside of any annotated region.
    \item \textbf{IN-REGION-TEMPORARY}: We are inside a annotated region that we started at the current nesting level.
    \item \textbf{IN-REGION-FORCED}: We are inside a annotated region that we started at a previous nesting level (e.g. we started it and then entered a group node subproblem).
\end{itemize}

Instead of a tuple of positions in the prototype and in target, we track a tuple of a position in the prototype and a ``confidence nesting level'', which represents how many ancestors of this node are in high-confidence regions rather than low-confidence regions. This allows us to keep track of how many group nodes we must exit before we are allowed to stop a low-confidence region.

\begin{sidewaysfigure}[t]
    \centering
    \includegraphics[width=\textwidth,trim=0.3in 10in 0.3in 0,clip]{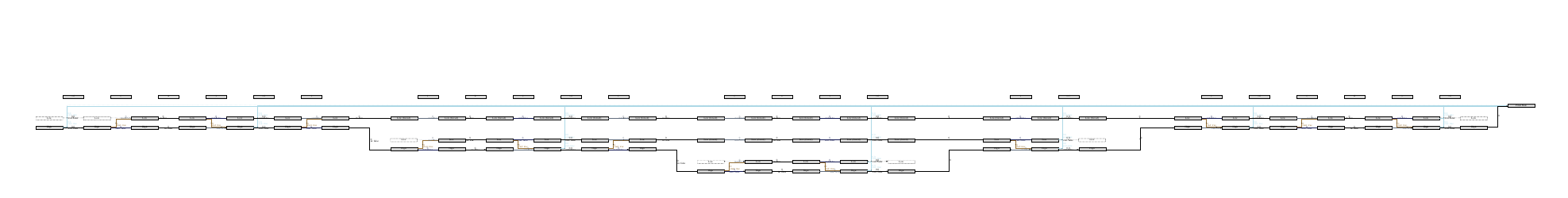}
    \caption{Rendering of the constraint decision diagram with all features enabled, for enforcing constraints in the sequence ``a [ b [ c ] ] d''.}
    \label{fig:app:constraint-dag}
\end{sidewaysfigure}

\cref{fig:app:constraint-dag} shows a rendering of the decision diagram we construct when combining two simple sequences. Note that the utility of this diagram is zero along any path; the purpose of this diagram is to forbid certain subsets of variable assignments (e.g assign them negative utility, or infinite cost).

\begin{figure}[t]
    \centering
    \includegraphics[width=\textwidth,trim=3.35in 10in 3.35in 0,clip]{figures/appendix/constraint-dag.pdf}
    \caption{Zoomed-in view of a portion of the constraint decision diagram in \cref{fig:app:constraint-dag}. ``Low'' refers to being in an annotated region, which corresponds to low-confidence \LowConf{} annotations for the edit localization task.}
    \label{fig:app:constraint-dag-zoom}
\end{figure}

\clearpage
\section{Overview of our Pseudo-Parser}
\label{app:pseudo-parser}

We now provide a high-level overview of our pseudo-parser, which converts code fragments into abstract syntax tree (AST) like structures. Pairs of these pseudo-parse trees are used by \RUSURE\ to construct matching graphs that parameterize the abstract space over which \RUSURE\ searches for minimum Bayes risk solutions. By representing the source code by a syntactically meaningful tree structure, it is possible for \RUSURE\ to produce completion results that respect the nature of source code and are especially syntactically meaningful.

Although a complete and precise specification of our pseudo-parsing algorithm is beyond the present scope, the full details will be available soon in our upcoming code release. In anticipation of that release, we now provide a high level description of our pseudo-parser and give some illustrative examples.

\subsection{High Level Desiderata}

We developed our bespoke pseudo-parser with two main goals in mind:
\begin{itemize}
    \item Language independence: our system handles \java , \javascript\ and \cpp\ using a unified approach that requires only a handful of language specific parameters. \python\ is handled similarly, but with some additional complexity due to nature of indents and dedents in that language.
    \item Error tolerance: we cannot assume syntactically valid code. 
\end{itemize}

\subsection{Algorithms}

\subparagraph{Tokenization} Before pseudo-parsing, we convert source code text into a sequence of tokens that we identify by regular expression matching. For example, the following code fragment:
\begin{verbatim}
y = func(x)
\end{verbatim}
is split into the following (token type, token) pairs:
\begin{verbatim}
ID           "y"
WHITE_SPACE  " "
PUNC         "="
WHITE_SPACE  " "
ID           "func"
BRACE        "("
ID           "x"
BRACE        ")"
NEWLINE      "\n"
\end{verbatim}

\subparagraph{Basic Bracket Matching}

Following tokenization comes the core part of our pseudo-parser, a \textit{bracket-matching} algorithm that produces a nested structure. For the example above, this may be rendered as below. While the full details of the following rendering are not important, the nesting denoted by indentation clearly reveals relevant structure:
\begin{verbatim}
GROUP(ROOT): "y = func(x)\n"
  GROUP(SPLIT_GROUP): "y = func(x)\n"
    TOK(CONTENT_LEAF): "y"
    DEC: " "
    TOK(CONTENT_LEAF): "="
    DEC: " "
    TOK(CONTENT_LEAF): "func"
    GROUP(MATCH): "(x)"
      TOK(MATCH_LEFT): "("
      GROUP(MATCH_INNER): "x"
        TOK(CONTENT_LEAF): "x"
      TOK(MATCH_RIGHT): ")"
    DEC: "\n"
\end{verbatim}

\subparagraph{Error Correction} How to handle sequences with unmatched brackets? In simple cases, missing closing brackets can be added to restore balance and recover relevant structure, for example this code:
\begin{verbatim}
(x
\end{verbatim}
results in the following tree structure:
\begin{verbatim}
GROUP(ROOT): "(x\n)\n"
  GROUP(SPLIT_GROUP): "(x\n)\n"
    GROUP(MATCH): "(x\n)"
      TOK(MATCH_LEFT): "("
      GROUP(MATCH_INNER): "x\n"
        TOK(CONTENT_LEAF): "x"
        DEC: "\n"
      TOK(MATCH_RIGHT): ")"
    DEC: "\n"
\end{verbatim}
which includes an additional closing brace.

\subparagraph{Error Tolerance} In some cases, such as the following: 
\begin{verbatim}
(x])
\end{verbatim}
no correction is made, and the erroneous brace (in this case the right square brace) is treated as a regular token, yielding
\begin{verbatim}
GROUP(ROOT): "(x])\n"
  GROUP(SPLIT_GROUP): "(x])\n"
    GROUP(MATCH): "(x])"
      TOK(MATCH_LEFT): "("
      GROUP(MATCH_INNER): "x]"
        TOK(CONTENT_LEAF): "x"
        TOK(CONTENT_LEAF): "]"
      TOK(MATCH_RIGHT): ")"
    DEC: "\n"
\end{verbatim}

\subparagraph{Handling \python\ indents and dedents} Unlike \cpp , \java\ and \javascript , which use curly brackets, the \python\ language uses white-space to denote code blocks. To handle this, we apply our pseudo parser twice. In the first step, we match standard brackets, so that 

\begin{verbatim}
def f(
  x, y):
  return x
\end{verbatim}
is parsed as
\begin{verbatim}
GROUP(ROOT): "def f(\n  x, y):\n  return x"
  TOK(CONTENT_LEAF): "def"
  DEC: " "
  TOK(CONTENT_LEAF): "f"
  GROUP(MATCH): "(\n  x, y)"
    TOK(MATCH_LEFT): "("
    GROUP(MATCH_INNER): "\n  x, y"
      DEC: "\n"
      DEC: "  "
      TOK(CONTENT_LEAF): "x"
      TOK(CONTENT_LEAF): ","
      DEC: " "
      TOK(CONTENT_LEAF): "y"
    TOK(MATCH_RIGHT): ")"
  TOK(CONTENT_LEAF): ":"
  DEC: "\n"
  DEC: "  "
  TOK(CONTENT_LEAF): "return"
  DEC: " "
  TOK(CONTENT_LEAF): "x"
\end{verbatim}
from which we determine (using a specific algorithm that works with the above tree representation), that since the newline and subsequent white-space following the opening bracket is contained within a matched bracket pair, it is not to be treated as a python code block indent. Once we have detected what do appear to be valid python code block indents and dedents, we handle them with a second pass of our error tolerant bracket matching pseudo-parser, which in this case gives the result:
\begin{verbatim}
GROUP(ROOT): "def f(\n  x, y):\n  return x\n"
  GROUP(SPLIT_GROUP): "def f(\n  x, y):\n  return x\n"
    GROUP(SPLIT_GROUP): "def f(\n  x, y):\n"
      TOK(CONTENT_LEAF): "def"
      DEC: " "
      TOK(CONTENT_LEAF): "f"
      GROUP(MATCH): "(\n  x, y)"
        TOK(MATCH_LEFT): "("
        GROUP(MATCH_INNER): "\n  x, y"
          DEC: "\n"
          DEC: "  "
          TOK(CONTENT_LEAF): "x"
          TOK(CONTENT_LEAF): ","
          DEC: " "
          TOK(CONTENT_LEAF): "y"
        TOK(MATCH_RIGHT): ")"
      TOK(CONTENT_LEAF): ":"
      DEC: "\n"
    GROUP(SPLIT_GROUP): "  return x\n"
      GROUP(MATCH): "  return x"
        TOK(MATCH_LEFT): ""
        GROUP(MATCH_INNER): "  return x"
          DEC: "  "
          TOK(CONTENT_LEAF): "return"
          DEC: " "
          TOK(CONTENT_LEAF): "x"
        TOK(MATCH_RIGHT): ""
      DEC: "\n"
\end{verbatim}
in which the matching python indents and dedents are denoted by empty strings.

\subparagraph{Subtokenization of string literals} To allow fine-grained edits within strings (such as docstrings), we further subtokenize tokens identified as string literals. This subtokenization process uses a generic lossless tokenizer originally designed by \citet{Kanade2019LearningAE} and made available at \url{https://github.com/google-research/google-research/tree/master/cubert/unified_tokenizer.py}.
\clearpage
\section{Additional Details on the Experimental Methodology}
\label{app:experiment-details}

\subsection{Example Generation}

In this section we provide further details on the example generation methodology introduced in \cref{sec:experiments:localizing}. An example is defined by
\begin{enumerate}
    \item The choice of source code file from which to derive the example. We use permissively licensed code from scientific computing repositories hosted on \textsc{Github}\footnote{\url{https://github.com}}.
    \item The starting (or cursor) location at which the hypothetical completion should begin, which is an index into the characters of the raw source code file.
    \item The truncation point, which is the assumed ending location of both the ground truth target (taken from the original source file, and used for evaluation but not seen by \RUSURE ), and each of the $K=31$ continuations that are samples drawn from the language model (and are used to form the minimum Bayes risk objective of \RUSURE\ defined in \cref{eqn:objective-finite-sample}).
\end{enumerate}

The first two example types, applicable to all four programming languages that we consider, choose the starting location uniformly at random from the source code file, and only differ by their choice of truncation point as follows. 
\begin{enumerate}
    \item For the \textbf{untruncated target} setting, we simply let the truncation point be the end of the source code file (or the maximum number of tokens allowed in the model's completion).
    \item For the \textbf{pseudo-parser heuristic target} method, we attempt to construct an evaluation target that is more tailored to practical settings, by truncating at a heuristically defined point beyond which further continuation may be overly ambiguous. To this end, we first pseudo parse the example code without truncation, and then find the nearest location following the starting (or cursor) position which either i) corresponds to the end of the nested sub-tree which contains the starting location (roughly speaking, the end of the current curly-braced block in \java , say), or if this does not exist because the cursor was not within a nested part of the source code file, ii) terminates at the end of the current statement (roughly speaking, the next semi-colon in \java , say).
\end{enumerate}

The final \textbf{pydocstring target} example type is \python\ specific and designed to yield a different distribution of examples that include a significant natural language component. To achieve this we let the starting location of the example be the beginning of a \textsc{DocString} comment in the source code file (immediately after the triple quotes) and the truncation point be immediately after the corresponding closing triple quotes. To identify the \textsc{DocString}, we again lean on our pseudo-parser: we search for triple quotes that occur at the beginning of indented code blocks. For the model samples, there is no guarantee that the \textsc{DocString} will be correctly closed; in such cases we simply fall back to the untruncated target approach.

\subparagraph{Removing the context} Finally, we note that due to our dataset construction strategy, and inspired by real-world code completion systems, our suggestions may begin partway in the middle of an expression. We address this by concatenating the context (the prefix of the file) and the model suggestions, pseudo-parsing the result, and then removing any node that is entirely contained in the context after parsing. This enables us to build a tree representation of only the part of the code that we would actually be suggesting, while still having its tree structure match the parse tree of the final code state.

\subsection{Utility function configuration}\label{app:experiment-details-utility-configuration}
We configure our base utility function (described in \cref{app:utility-functions}) in different ways for each task.

\paragraph{Edit localization task.} For this task, we configure the utility function with a per-character utility of 1 per matched \HighConf{} token and $\alpha = 0.7$ per matched \LowConf{}, and a per-character cost of 1 per deleted \HighConf{} and $\beta = 0.3$ per deleted \LowConf{}; this setting is such that tokens with a lower-than-70\% chance of being kept are optimal to mark as \LowConf{}. (We vary these thresholds for the Pareto plot, by setting the \LowConf{} match utility to $\alpha = c$ and deletion cost to $\beta = 1 - c$ for varying $c$.) We also include a localization penalty of 5 per edit inside \HighConf{} regions, a penalty of 0.25 in \LowConf{} regions, and a penalty of 0.75 for starting a new \LowConf{}. These costs are also tuned so that, if there is a 30\%-or-greater chance of starting to edit at a given location, it is better to insert a \LowConf{} region that includes the edit.

\paragraph{Prefix task.} We use the same configuration as the edit localization task, but additionally insert truncation nodes into the prototype suggestion, which enables us to search over points to stop the suggestion early. For the R-U-SURE (Prefix) variant, we do not insert any Region Start / Region End nodes, which forces the solver to label everything as \HighConf{} and only search for prefixes. For the R-U-SURE (Prefix + Region) variant, we include both Region Start/End nodes and truncation nodes.

\paragraph{API call task.} For this task, we restrict our attention to Python files, and do additional postprocessing on both the model samples and the ground truth target in order to compute an estimated utility. We first search through the parsed file in order to identify statements that look like function calls; in particular, any statement that contains tokens immediately followed by an open parenthesis, and which does not start with \texttt{def} or \texttt{class}. We then extract a list of such calls and rearrange them into a shallow tree structure: the top level sequence is a sequence of group nodes, and each group node contains exactly one call. We further insert region start/end nodes into each call, before and after the parenthesis, respectively; these allow our method to decide how many attribute accesses to include in the call (e.g. `foo.bar.baz(` or just `bar.baz`) and whether or not to include arguments or a left-hand-side assignment. For this task, we reinterpet the regions as being \HighConf{} rather than \LowConf{}. Since we only care about extracting a useful subsequence, we forbid any token matches outside of extracted regions, but set the costs of deletion and insertion to zero. We also forbid any deletions or insertions in an extracted region to ensure that the call matches exactly (instead of just having high token overlap). We implement this by building a simplified version of our edit distance graph that only includes nodes for the allowed types of edit.

Within an extracted region, we compute per-token weights, which are 1 for tokens we have already seen in the file and 10 for novel tokens (those not yet seen in the file); we also give 1 bonus point for correctly predicting the entire argument list. We then scale this base weight by 0.7 to get the utility of correct predictions, and scale it by 0.3 to get the penalty for incorrect ones. Note that this is the same set of rewards and penalties as in the edit localization task, however, the break-even point is lower for this version because deleting tokens has a penalty of 0 instead of -1.

\clearpage
\section{Detailed Experimental Results}\label{app:detailed-experimental-results}

\begin{figure*}[h!]
    \centering
     \begin{subfigure}[b]{0.35\textwidth}
         \centering
         \includegraphics[width=\textwidth,page=1]{figures/results/Regions_5000_64_samples_logits_metrics_UncertaintyRegionsWrapper_288_size_ablation_do_subtract_baseline_maybe-crop.pdf}
         \caption{Ground Truth.}
         \label{fig:sizeablation:gt}
     \end{subfigure}
     \begin{subfigure}[b]{0.3551\textwidth}
         \centering
         \includegraphics[width=\textwidth,page=2]{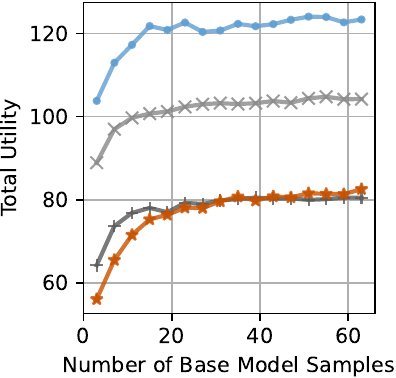}
         \caption{Leave one out.}
         \label{fig:sizeablation:loo}
     \end{subfigure}
    \caption{
     The dependence of model performance on the number of base model samples combined by \RUSURE , evaluated with respect to the ground truth user intent (left) and a leave one out sample from the base model (right). The four lines represent the four programming languages we considered. We observe that the performance increases dramatically on the left, but that this increase is relatively flat around our maximum considered \MAXNUMBEROFMODELSAMPLES\ samples.
    }
    \label{fig:appendix:sizeablation}
\end{figure*}

The additional detailed results provided in the appendix include:
\begin{itemize}
    \item \cref{fig:rusuregionsbarplot}: A detailed breakdown by target type (\textsc{Untruncated}, \textsc{Heuristic} and \textsc{PyDocstring}) for both leave one out and ground truth targets, of the performance of the \RUSUREREGIONS\ method.
    \item \cref{fig:dualitygap:regions}: An analysis of the duality gap achieved by our dual decomposition solver, that shows that the optimal solution is found in the majority of cases.
    \item \cref{fig:appendix:sizeablation}: A plot of model performance by size of sample $K$ which includes both leave one out and ground truth utilities.
    \item \cref{tab:appendix:uncertainty-regions}, \cref{tab:appendix:prefix-length} and
    \cref{tab:appendix:api-discovery}: Detailed versions of the tables in the main paper.
\end{itemize}

\clearpage

\newcommand\barplotwidth{0.4}
\begin{figure*}[h!]
    \centering
    \includegraphics[width=\barplotwidth\textwidth,page=1]{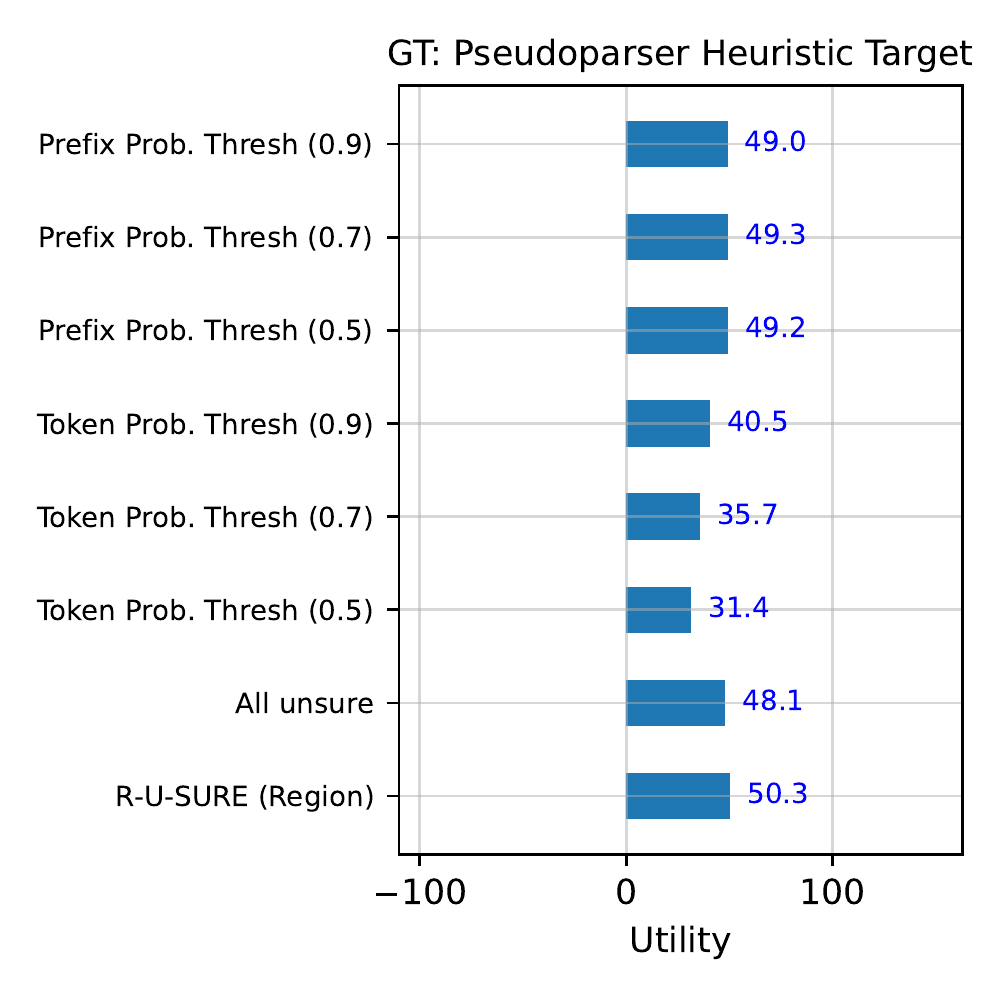}
    \includegraphics[width=\barplotwidth\textwidth,page=2]{figures/results/Regions_5000_32_samples_logits_metrics_UncertaintyRegionsWrapper_144_barplot.pdf}
    \includegraphics[width=\barplotwidth\textwidth,page=3]{figures/results/Regions_5000_32_samples_logits_metrics_UncertaintyRegionsWrapper_144_barplot.pdf}
    \includegraphics[width=\barplotwidth\textwidth,page=4]{figures/results/Regions_5000_32_samples_logits_metrics_UncertaintyRegionsWrapper_144_barplot.pdf}
    \includegraphics[width=\barplotwidth\textwidth,page=5]{figures/results/Regions_5000_32_samples_logits_metrics_UncertaintyRegionsWrapper_144_barplot.pdf}
    \includegraphics[width=\barplotwidth\textwidth,page=6]{figures/results/Regions_5000_32_samples_logits_metrics_UncertaintyRegionsWrapper_144_barplot.pdf}
    \caption{
    Average utility (higher is better) for our \RUSUREREGIONS\ and a variety of baseline methods for the uncertainty-regions task, evaluated on the ground truth (GT, left) user intent as well as a leave-one-out (LOO, right) sample from the model, and the three target settings (corresponding to the three rows of plots) noted in the figure titles. Note that methods that perform well in the leave-one-out setting also tend to perform well on the ground truth, but averages are slightly better across the board for leave-one-out.
    }
    \label{fig:rusuregionsbarplot}
\end{figure*}
\newcommand\dualitygapwidth{0.4}
\begin{figure*}[h!]
    \centering
     \begin{subfigure}[b]{\dualitygapwidth\textwidth}
         \centering
         \hspace{-15mm}\includegraphics[width=\textwidth]{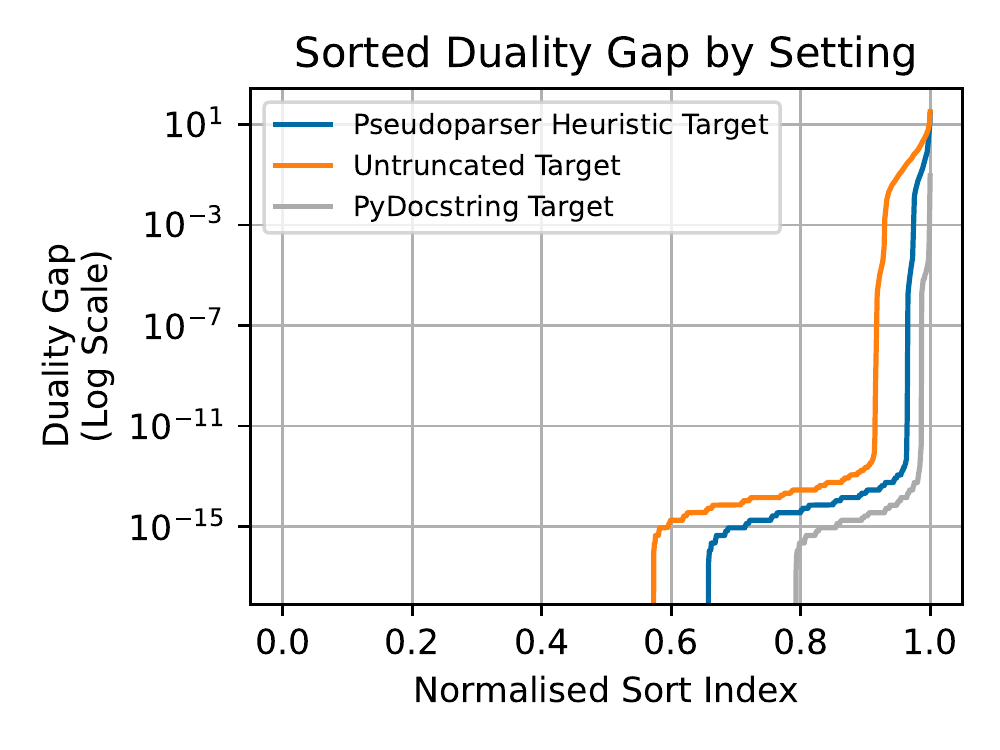}
         \caption{\RUSUREREGIONS .}
         \label{fig:dualitygap:regions}
     \end{subfigure}
     \begin{subfigure}[b]{\dualitygapwidth\textwidth}
         \centering
         \hspace{-15mm}\includegraphics[width=\textwidth]{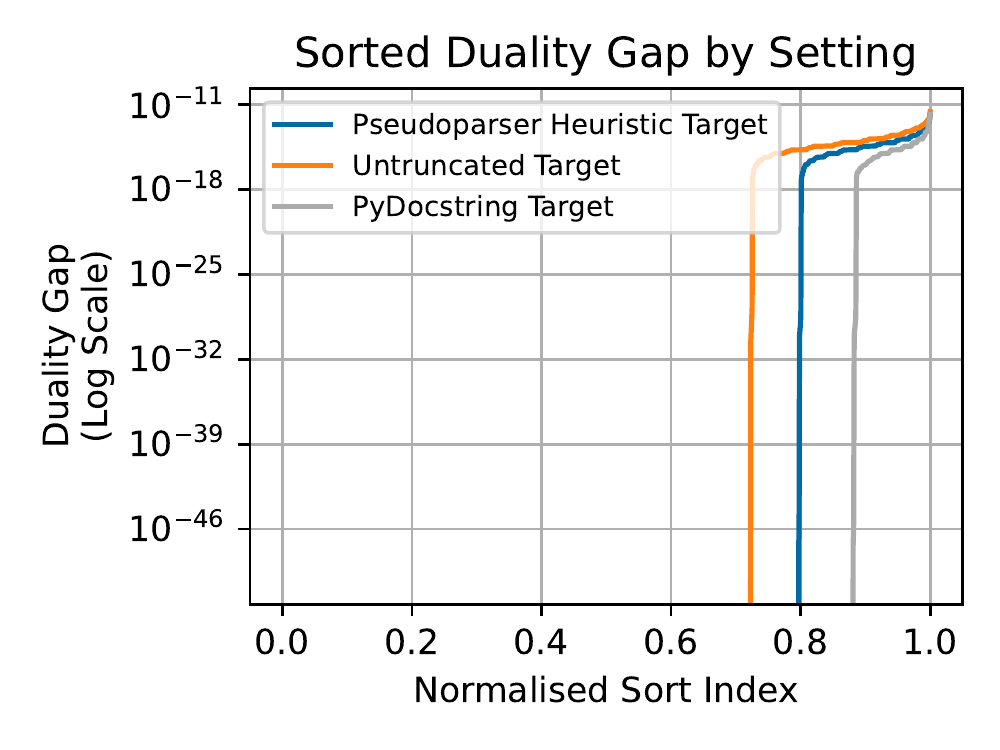}
         \caption{\RUSUREPREFIX .}
         \label{fig:dualitygap:prefix}
     \end{subfigure}
     \begin{subfigure}[b]{\dualitygapwidth\textwidth}
         \centering
         \hspace{-15mm}\includegraphics[width=\textwidth]{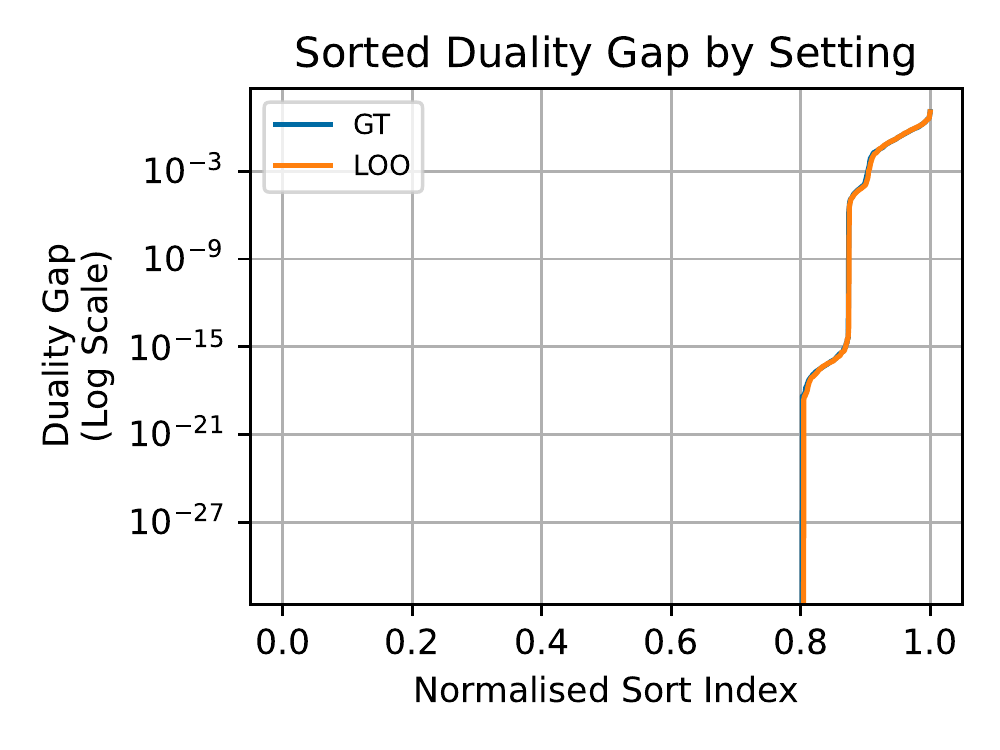}
         \caption{\RUSUREAPI .}
         \label{fig:dualitygap:api}
     \end{subfigure}
    \caption{
    The distribution of duality gaps presented as a log plot of the sorted values, broken down by utility function for each figure (a)-(c), and with a separate line for each type of prediction target (for (a) and (b)) or prediction target type (i.e. ground-truth or leave one out, for (c)). We observe that \RUSUREPREFIX\ always obtains practically zero gap (and hence primal optimality), while \textit{e.g.} \RUSUREREGIONS\ does so on around 90-98\% of cases, depending on the type of prediction target.
    }
    \label{fig:appendix:dualitygap}
\end{figure*}
\begin{table*}[t]
    \centering
{\footnotesize
\begin{tabular}{ccccccccc}
\toprule
{}  & \thead{Utility \\ {\scriptsize (relative)}} & \thead{Est. Utility \\ {\scriptsize (relative)}} & \thead{LOO Utility \\ {\scriptsize (relative)}} & \thead{Sensitivity \\ {\scriptsize \% \LowConf{} of edited}} & \thead{Specificity \\ {\scriptsize\% \scriptsize \HighConf{} of unedited}} & $F_1$ score \\
\midrule
{\textsc{All Sure}}
&
$\equiv 0$ &
38.00 &
30.35 &
0.00 &
100.00 &
-
\\
{\textsc{Maximal Unsure}}        &             81.83 &              106.08 &             101.90 &                            90.79 &                             6.85 & 12.74 \\
{\textsc{Token Prob. 0.5}}  &             50.83 &               82.63 &              77.10 &                            55.63 &                            72.69 & 63.03  \\
{\textsc{Token Prob. 0.7}}  &             58.42 &               88.89 &              83.68 &                            63.81 &                            64.97 & 64.38 \\
{\textsc{Token Prob. 0.9}}  &             66.99 &               95.64 &              90.79 &                            73.21 &                            52.93 & 61.44 \\
{\textsc{Prefix Prob. 0.5}} &             83.33 &              108.52 &             104.29 &                            89.31 &                            27.39 & 41.92 \\
{\textsc{Prefix Prob. 0.7}} &             83.45 &              108.27 &             104.05 &                            89.89 &                            23.50 & 37.26 \\
{\textsc{Prefix Prob. 0.9}} &             83.08 &              107.61 &             103.41 &                            90.35 &                            17.65 & 29.53\\
{\textsc{Ours} (Region)}         &             \textbf{84.42} &              \textbf{113.82} &             \textbf{109.12} &                            85.78 &                            62.24 & \textbf{72.14} \\
\bottomrule
\end{tabular}
}

\caption{Detailed results for our \RUSUREREGIONS\ method along with a selection of baselines, on the edit-localization task.}
    \label{tab:appendix:uncertainty-regions}
\end{table*}
\begin{table}[t!]
    \centering
    \footnotesize{
\begin{tabular}{crrrrr}
\toprule
{} & \thead{GT Utility \\ (mean)} & \thead{Est. Utility \\ (mean)} & \thead{LOO \\ Utility (mean)} & \thead{Correct Chars} & \thead{Incorrect Chars} \\
\midrule
\textsc{20 Characters} &-7.99 &5.15 &4.11 &13.61 &21.60 \\
\textsc{50 Characters} &-8.92 &7.12 &5.55 &23.83 &32.75 \\
\textsc{100 Characters} &-14.75 &4.96 &2.47 &34.35 &49.10 \\
\textsc{200 Characters} &-29.88 &-5.04 &-9.20 &44.81 &74.69 \\
\textsc{500 Characters} &-66.59 &-33.18 &-40.16 &53.66 &120.25 \\
\midrule
\textsc{1 Line} &-10.91 &4.00 &2.45 &16.85 &27.76 \\
\textsc{2 Lines} &-12.68 &4.85 &2.88 &24.30 &36.98 \\
\textsc{4 Lines} &-19.85 &1.76 &-1.04 &33.75 &53.59 \\
\textsc{8 Lines} &-37.75 &-9.77 &-14.31 &43.97 &81.72 \\
\textsc{16 Lines} &-65.76 &-31.32 &-38.14 &51.82 &117.58 \\
\midrule
\textsc{Token Prob. 0.00} &-84.46 &-47.36 &-55.52 &54.94 &139.40 \\
\textsc{Token Prob. 0.01} &-13.46 &91.15 &6.31 &38.85 &52.32 \\
\textsc{Token Prob. 0.02} &-7.56 &78.02 &10.78 &35.24 &42.80 \\
\textsc{Token Prob. 0.05} &-2.80 &63.21 &13.56 &30.21 &33.01 \\
\textsc{Token Prob. 0.10} &-0.58 &53.57 &14.44 &26.50 &27.08 \\
\textsc{Token Prob. 0.20} &0.52 &44.77 &13.09 &22.65 &22.13 \\
\textsc{Token Prob. 0.30} &0.69 &14.35 &13.43 &20.45 &19.76 \\
\textsc{Token Prob. 0.50} &0.19 &12.82 &12.07 &17.30 &17.12 \\
\textsc{Token Prob. 0.70} &-1.10 &10.72 &10.01 &14.42 &15.52 \\
\textsc{Token Prob. 0.90} &-3.80 &7.21 &6.54 &10.62 &14.42 \\
\midrule
\textsc{Prefix Prob.0.01} &0.88 &48.99 &15.50 &24.94 &24.06 \\
\textsc{Prefix Prob.0.02} &1.04 &46.30 &15.25 &23.68 &22.64 \\
\textsc{Prefix Prob.0.05} &1.03 &42.62 &14.64 &21.83 &20.80 \\
\textsc{Prefix Prob.0.10} &0.83 &39.57 &13.99 &20.20 &19.37 \\
\textsc{Prefix Prob.0.20} &0.43 &36.28 &14.36 &18.36 &17.93 \\
\textsc{Prefix Prob.0.30} &0.04 &34.01 &12.26 &17.03 &16.99 \\
\textsc{Prefix Prob.0.50} &-1.00 &30.50 &10.63 &14.76 &15.75 \\
\textsc{Prefix Prob.0.70} &-2.40 &27.39 &8.58 &12.50 &14.90 \\
\textsc{Prefix Prob.0.90} &-5.01 &23.43 &5.52 &9.21 &14.22 \\
\midrule
\textsc{Max Avg. Log Prob} &-17.64 &50.96 &-4.08 &16.66 &34.31 \\
\textsc{Intellicode Compose} &0.04 &12.71 &11.90 &17.10 &17.06 \\
\textsc{Ours (Prefix)} &\textbf{7.00} &\textbf{30.49} &\textbf{28.03} &38.81 &31.81 \\
\midrule
\textsc{Ours (Prefix+Region)} &\textbf{12.26} &\textbf{37.79} &\textbf{35.18} &36.40 &22.31 \\
\bottomrule
\end{tabular}
    }
    \caption{
    Comparison of utility and character-level accuracy statistics for the suggestion-length task; \RUSUREPREFIX{} achieves higher average utility than the comparable baselines. As an additional comparison, we include results for \RUSUREPREFIXPLUSUNSURE{}, a variant that is also allowed to mark some tokens \LowConf{}, which improves our utility metric and decreases the number of incorrectly-predicted characters (where we only count \HighConf{} tokens as correct/incorrect).
    }
    \label{tab:appendix:prefix-length}
\end{table}

\begin{table}[t!]
    \centering
\begin{tabular}{crrrrrrrrr}
\toprule
{} & \scriptsize{\thead{GT Utility\\(mean)}} & \scriptsize{\thead{Est. Utility \\ (mean)}} & \scriptsize{\thead{LOO Utility \\ (mean)}} & \scriptsize{\thead{Corr. \\ (total)}} & \scriptsize{\thead{ Corr. \\ (novel)}} & \scriptsize{\thead{Corr. \\ (not novel)}} & \scriptsize{\thead{Incorr. \\ (total)}} & \scriptsize{\thead{Incorr. \\ (novel)}} & \scriptsize{\thead{Incorr. \\ (not novel)}} \\
\midrule
\footnotesize{\textsc{All Full}}      
&-8.75 &-6.09 &-7.15 &2.02 &0.33 &1.68 &9.64 &3.28 &6.36 \\
\footnotesize{\textsc{All Short}}   
&-0.58 &0.70 &0.10 &3.00 &0.68 &2.32 &4.60 &2.07 &2.53 \\
\footnotesize{\textsc{Novel Short}} 
&-1.39 &-0.36 &-0.94 &1.18 &0.68 &0.50 &3.04 &2.07 &0.96 \\
\footnotesize{\textsc{Ours (API)}}        
&\textbf{5.10} &\textbf{6.74} &\textbf{6.53} &3.56 &0.68 &2.88 &2.10 &0.50 &1.60 \\
\bottomrule 
\end{tabular}
    \caption{
    Detailed results for our \RUSUREAPI\ method along with a set of baselines.}
    \label{tab:appendix:api-discovery}
\end{table}

\end{document}